\documentclass[lettersize,journal]{IEEEtran}
\usepackage{amsmath,amsfonts}
\usepackage{algorithmic}
\usepackage{algorithm}
\usepackage{array}
\usepackage[caption=false,font=normalsize,labelfont=sf,textfont=sf]{subfig}
\usepackage{textcomp}
\usepackage{stfloats}
\usepackage{url}
\usepackage{verbatim}
\usepackage{graphicx}
\usepackage{cite}
\usepackage{tcolorbox}
\usepackage{enumitem} 
\usepackage{booktabs}
\usepackage{graphicx}
\usepackage{xcolor}
\usepackage{colortbl}

\usepackage{booktabs}
\usepackage{multirow}
\usepackage{xcolor}
\usepackage{makecell}

\usepackage{arydshln}
\usepackage[table]{xcolor}

\usepackage{caption}  

\newcommand\our{\textsc{ScenePerson-13W}}

\begin{document}

\title{SA-Person: Text-Based Person Retrieval with Scene-aware Re-ranking}


\author{Yingjia~Xu$^{*}$, Jinlin~Wu$^{*}$, Daming~Gao, Zhen~Chen, Yang~Yang, \\
 Min~Cao, Mang Ye,~\IEEEmembership{Senior Member,~IEEE} and Zhen~Lei,~\IEEEmembership{Fellow,~IEEE}%


        
\thanks{
Yingjia Xu, Daming Gao and Min Cao are with the School of Computer Science and Technology, Soochow University, Suzhou 215006, China (e-mail: yjxuxuyj@stu.suda.edu.cn; daming\_gao@outlook.com, mcao@suda.edu.cn).\\
Jinlin Wu is with the Centre for Artificial Intelligence and Robotics, Hong Kong Institute of Science and Innovation, Chinese Academy of Sciences, Hong Kong, and also with the Multimodal Artificial Intelligence Systems (MAIS), Institute of Automation, Chinese Academy of Sciences, Beijing 100190, China (e-mail: jinlin.wu@nlpr.ia.ac.cn).\\
Zhen Chen, and Yang Yang are with the Centre for Artificial Intelligence and Robotics, Hong Kong Institute of Science and Innovation, Chinese Academy of Sciences, Hong Kong (e-mail: zhen.chen@yale.edu; yang.yang@nlpr.ia.ac.cn).\\
Mang Ye is with Wuhan University, Wuhan, China. 
(Email: yemang@whu.edu.cn). \\
Zhen Lei is with the Centre for Artificial Intelligence and Robotics, Hong Kong Institute of Science and Innovation, Chinese Academy of Sciences, Hong Kong, with the Multimodal Artificial Intelligence Systems (MAIS), Institute of Automation, Chinese Academy of Sciences, Beijing 100190, China, and also with the University of Chinese Academy of Sciences, Beijing 100049, China (e-mail: zlei@nlpr.ia.ac.cn).\\
\IEEEauthorblockA{\textsuperscript{*}Yingjia Xu and Jinlin Wu contributed equally to this work.\\
Corresponding author: Min Cao.}
}}


\maketitle

\begin{abstract}

Text-based person retrieval aims to identify a target individual from an image gallery using a natural language description.
Existing methods primarily focus on appearance-driven cross-modal retrieval, yet face significant challenges due to the visual complexity of scenes and the inherent ambiguity of textual descriptions.
The contextual information, such as landmarks and relational cues, provides complementary cues that can offer valuable complementary insights for retrieval, but remains underexploited in current approaches.
Motivated by this limitation, we propose a novel paradigm: scene-aware text-based person retrieval, which explicitly integrates both individual appearance and global scene context to improve retrieval accuracy.
To support this, we first introduce {\our}, a large-scale benchmark dataset comprising over 100,000 real-world scenes with rich annotations encompassing both pedestrian attributes and scene context.
Based on this dataset, we further present SA-Person, a two-stage retrieval framework.
In the first stage, SA-Person performs discriminative appearance grounding by aligning textual descriptions with pedestrian-specific regions.  
In the second stage, it introduces SceneRanker, a training-free, scene-aware re-ranking module that refines retrieval results by jointly reasoning over pedestrian appearance and the global scene context.
Extensive experiments on {\our} and existing benchmarks demonstrate the effectiveness of our proposed SA-Person. Both the dataset and code will be publicly released to facilitate future research.
\end{abstract}


\section{Introduction}
\label{introduction}

\IEEEPARstart{P}{erson} retrieval~\cite{galiyawala2021person} aims to identify a target individual within the image galleries based on a query, playing a vital role in applications like video surveillance and human-centric artificial intelligent systems~\cite{li2018richly,galiyawala2021person,li2023dcel}. 
Early efforts focus on image-based paradigms~\cite{zhang2016learning,liu2018pose,sarfraz2018pose,xiao2017joint,he2018end,yan2021anchor,cao2022pstr,yan2022exploring} using cropped images as queries, and then shift toward text-based retrieval~\cite{li2017person,yu2019cross,liu2023survey} due to the accessibility of textual descriptions.
Advances in large-scale Vision-Language Pretraining (VLP) models, such as CLIP~\cite{radford2021learning}, also accelerate the development of text-based person retrieval~\cite{yang2023towards,li2024adaptive,zuo2024ufinebench,luo2025graph,liu2025dm,jiang2023cross,bai2023rasa,cao2024empirical,ergasti2024mars,qin2024noisy}.
However, methods for text-based person retrieval rely solely on cropped images and appearance-focused textual descriptions, limiting their effectiveness in real-world scenarios. 
Recent methods~\cite{zhang2023text,su2024maca,yan2025fusionsegreid} extend retrieval to full-scene images but still focus narrowly on appearance, neglecting critical scene cues. 

\begin{figure*}[t]
    \centering
    \includegraphics[width=1\linewidth]{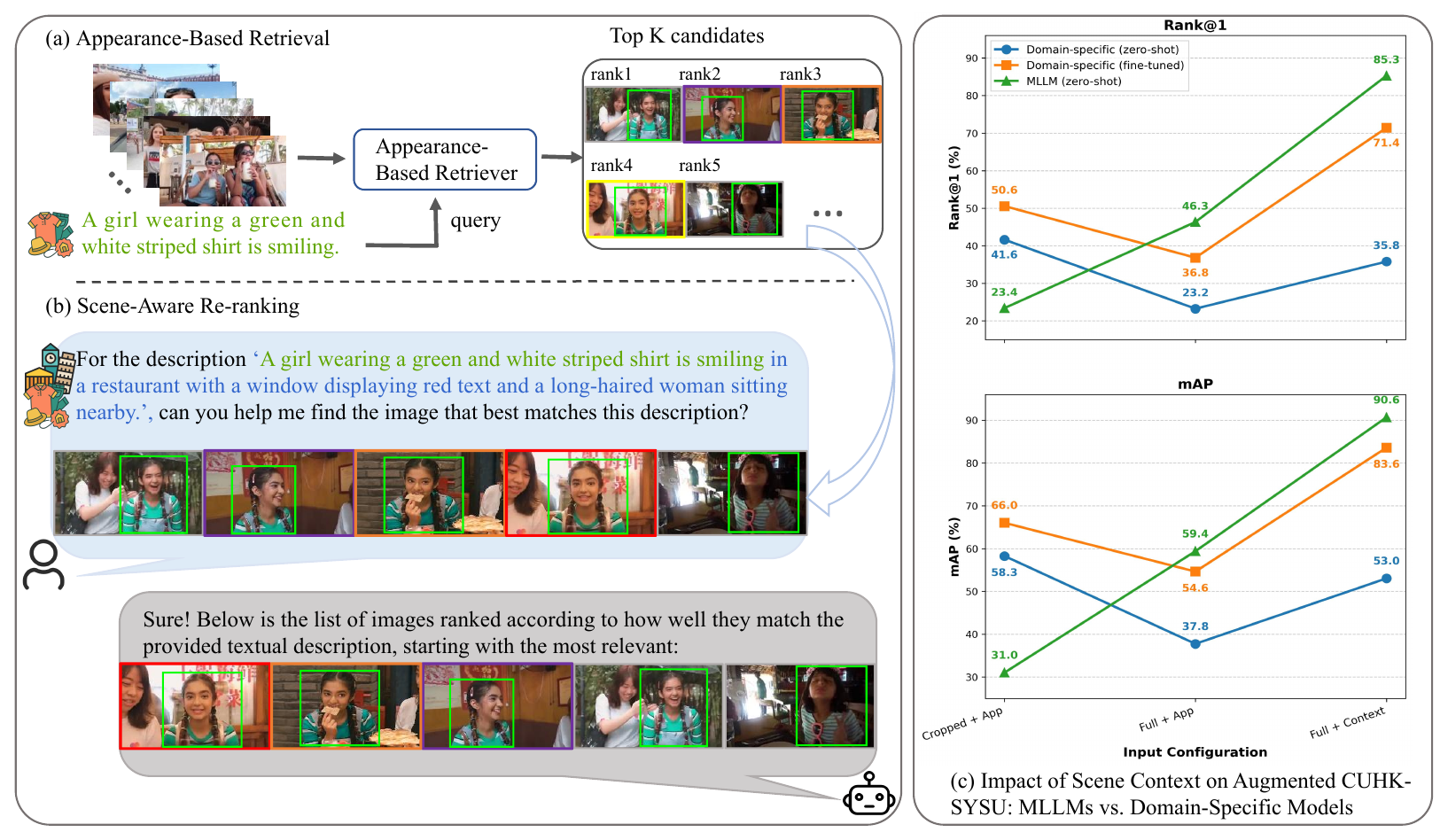}
    \caption{
    Overview of person retrieval challenges and scene-aware insights.
    (a) Illustration of appearance-based retrieval limitations, where multiple candidates match the text \textit{A girl wearing a green and white striped shirt is smiling} due to similar appearances. 
    (b) Demonstration of scene-aware reranking, ranking candidate images by aligning scene context like \textit{located in a restaurant, near a window with red text} with the text. 
    (c)
    Retrieval performance across three progressive input configurations on the augmented CUHK-SYSU dataset. 
    The three configurations are: (1) \textit{Cropped + App} (cropped image with appearance-only text), (2) \textit{Full + App} (full image with appearance-only text), and (3) \textit{Full + Context} (full image with context-enriched text). 
    }
    \label{fig:teaser}
\end{figure*}

This limitation poses challenges in complex scenarios with similar-looking individuals, or the same person under varying contexts, as illustrated in Figure~\ref{fig:teaser}(a).
In such cases, appearance-only retrieval often produces ambiguous or incorrect matches. 
By contrast, human descriptions about a target person are rarely restricted to appearance; instead, they naturally incorporate auxiliary contextual cues that complement visual traits, 
such as surrounding environments, landmark buildings, and co-occurring entities.
As illustrated in Figure~\ref{fig:teaser}(b), incorporating scene context substantially enhance disambiguation beyond what appearance features alone can offer. 
To quantify the potential of scene context, we perform a preliminary analysis on the 
CUHK-SYSU dataset augmented with MLLM-generated descriptions. As shown in 
Figure~\ref{fig:teaser}(c), MLLMs with context-enriched text can achieve retrieval performance beyond the limits of fine-tuned domain-specific models.
A comprehensive discussion is provided in Section~\ref{sec:Preliminary Analysis}.
Motivated by this observation, we focus on scene-aware text-based person retrieval, which explicitly incorporates both individual appearance and global scene context for more accurate and robust retrieval.

The first challenge is the lack of a dedicated dataset for this task. 
To this end, we introduce {\our}, a novel dataset designed for real-world scene-aware person retrieval. It includes over $131,438$ images from $12,562$ YouTube video sequences, featuring diverse individuals across varying crowd densities, camera angles, and environments. As summarized in Table~\ref{tab:dataset_comparison}, existing datasets\cite{zheng2015scalable,li2014deepreid,xiao2017joint,zheng2017person,li2017person,ding2021semantically,zhang2023text} typically focus on cropped images or weakly grounded captions, {\our} provides rich image-text pairs with detailed descriptions encompassing both visual appearance and scene context, which promotes text-to-person alignment. 

Building on {\our}, we propose SA-Person, a two-stage framework for scene-aware person retrieval. In the first stage, appearance-related information is extracted from textual descriptions to identify candidate pedestrians via cross-modal matching, reducing the retrieval space with discriminative appearance features. 
In the second stage, we introduce SceneRanker, a training-free, scene-aware re-ranking module that refines retrieval results by jointly reasoning over pedestrian appearance and global scene context, enabling holistic understanding to effectively disambiguate visually similar individuals and significantly improve retrieval performance.

In summary, our main contributions are as follows:
\begin{enumerate}[label=\arabic*)]

    \item We introduce \textbf{\our}, the first dataset enabling a new retrieval paradigm of scene-aware text-based person retrieval by providing compositional descriptions that include both pedestrian appearance and global scene context, alleviating the limitations of prior datasets.

    \item
    We present \textbf{SA-Person}, a two-stage retrieval framework that combines discriminative appearance grounding with holistic scene understanding, enabling robust retrieval from large-scale full-scene image databases.
    \item 
    We propose \textbf{SceneRanker}, the first re-ranking strategy for text-based person retrieval that can be seamlessly integrated as a plug-and-play module, incorporating visual grounding and contextual reasoning to disambiguate and enhanace retrieval performance in complex scenes.
    \item Extensive experiments on {\our} and diverse benchmarks demonstrate the effectiveness and generalizability of our proposed SA-Person across various retrieval settings and model backbones.
\end{enumerate}

\begin{figure*}[ht]
    \centering
    \includegraphics[width=1\textwidth]{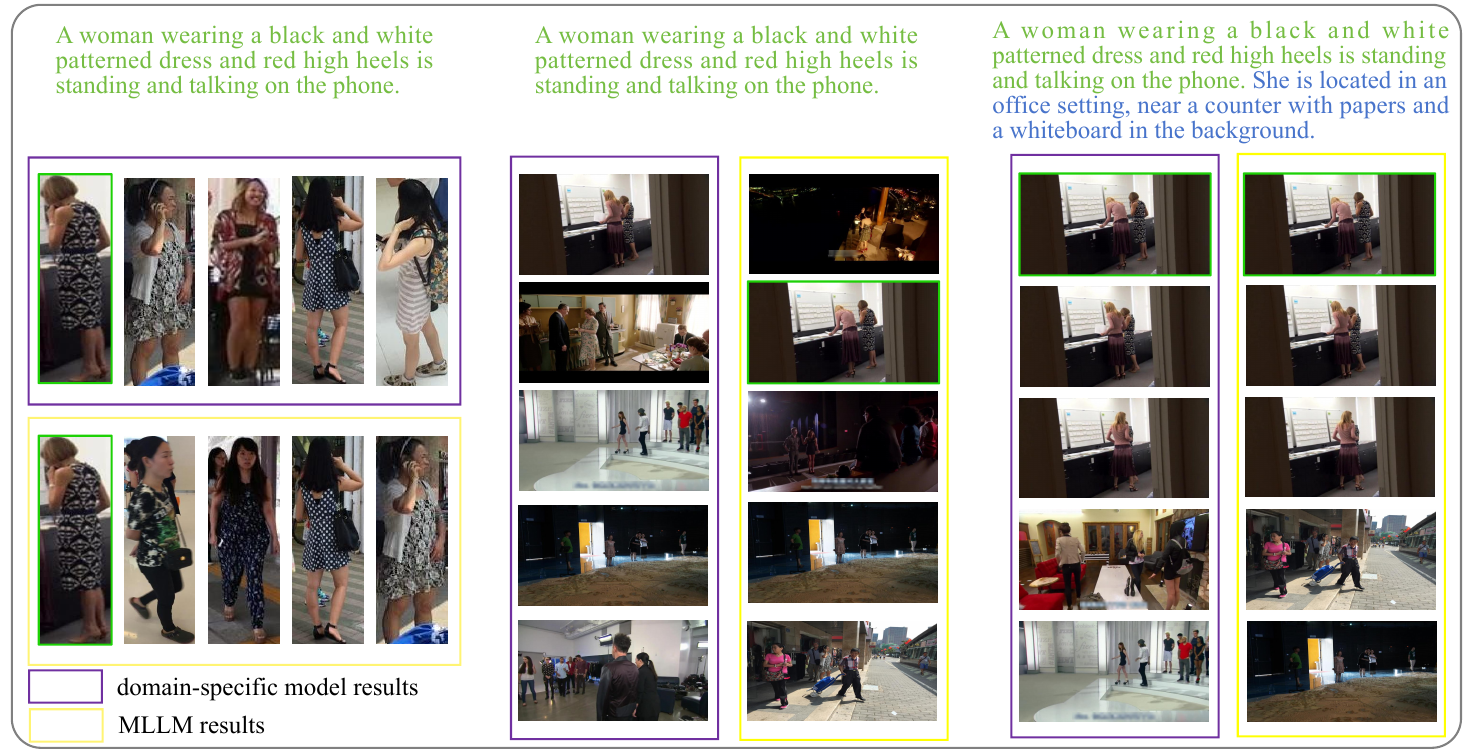} 
    \caption{
Qualitative visualization of retrieval results across three progressive input configurations on the augmented CUHK-SYSU dataset. 
    The blue frame contains the top candidates from the domain-specific model IRRA (ViT-L/14), and the orange frame contains the top candidates from the MLLM (InternVL-8B).
    The green-bordered image is the ground truth.
    }
    \label{fig:retrieval_comparison_preliminary}
\end{figure*}

    


\section{Related works}
\subsection{Text-Based Person Retrieval}
Text-Based Person Retrieval (TBPR) aims to retrieve specific individuals based on natural language descriptions, typically from a gallery of cropped person images~\cite{yang2023towards,li2024adaptive,zuo2024ufinebench,luo2025graph,liu2025dm}.
These methods focus on fine-grained cross-modal alignment between visual appearances and textual attributes, such as clothing, gender, or accessories~\cite{shu2022see,zuo2024ufinebench,wang2024fine,wei2024fine,zhao2025cross}.
With the advent of vision-language models like CLIP~\cite{radford2021learning}, TBPR performance has significantly improved. 
For instance, IRRA~\cite{jiang2023cross} introduces a multi-modal
interaction encoder for implicit relational reasoning, and TBPS-CLIP~\cite{cao2024empirical} proposed a simple yet strong baseline through enhanced training strategies.
To address noise in cross-modal data,  RaSa~\cite{bai2023rasa} proposed sensitivity-aware and relation-aware representation learning for robustness against weakly matched pairs;
MARS~\cite{ergasti2024mars} incorporated visual reconstruction and attribute-sensitive losses to mitigate inter-identity noise and intra-identity variations; RDE~\cite{qin2024noisy} introduced dual embeddings to handle under-correlated or mislabeled image-text pairs under noisy conditions.
Beyond cropped-image retrieval, some methods~\cite{zhang2023text,su2024maca,yan2025fusionsegreid} retrieval individuals directly from unprocessed full-scene images,
where individuals must be located and identified in complex visual environments 
Specifically, Zhang et al.~\cite{zhang2023text} integrated textual cues into region proposals, enabling text descriptions to guide the detection process.
Su et al.~\cite{su2024maca} introduced a memory-aided coarse-to-fine alignment strategy, using memory banks to enhance cross-modal contrastive learning.
Yan et al.~\cite{yan2025fusionsegreid} explored multimodal fusion by combining textual and visual features with precise segmentation techniques to boost retrieval performance.
While these methods often overlook the rich contextual information present in the surrounding scene, which could offer valuable clues for more accurate and robust person retrieval.

\subsection{Multimodal Large Language Models for Scene and Region-Level Reasoning}
Recent advances in Multimodal Large Language Models (MLLMs) enabled powerful visual reasoning across complex scenes. Current efforts primarily progresses along two directions: comprehensive scene understanding using global image-text inputs~\cite{chen2024sharegpt4v,chen2024expanding,bai2025qwen2,touvron2023llama,mitra2024compositional,fan2024mllm}, and fine-grained region-level grounding~\cite{guo2024regiongpt,zhou2023regionblip,peng2023kosmos,chen2023shikra,ma2024groma,rasheed2024glamm}.
For comprehensive scene understanding, recent works~\cite{mitra2024compositional,fan2024mllm} enables MLLMs to explicitly represent structured and compositional semantics. For instance, Mitra et al. ~\cite{mitra2024compositional} proposed generating scene graphs via MLLMs and incorporating them into prompts, facilitating compositional reasoning without relying on annotated scene graphs; MLLM-SUL ~\cite{fan2024mllm} introduced a dual-branch encoder and a fine-tuned LLaMA~\cite{touvron2023llama} model to jointly perform scene description in traffic scenarios.
To support region-level grounding, recent studies ~\cite{guo2024regiongpt,zhou2023regionblip,chen2023shikra} enhance MLLMs with explicit spatial conditioning. For example, RegionGPT~\cite{guo2024regiongpt} and RegionBLIP~\cite{zhou2023regionblip} have been proposed adopt region-wise feature pooling within bounding boxes or masks to extract localized representations for improved spatial reasoning; 
KOSMOS-2~\cite{peng2023kosmos} and Shikra~\cite{chen2023shikra} encode spatial positions as discrete tokens or numeric coordinates to support referential understanding; GLaMM~\cite{rasheed2024glamm} enables grounded interaction through joint generation of text and segmentation masks.

In this work, we combine the complementary strengths of domain-specific TBPR models and MLLMs for scene-aware text-based person retrieval, fusing efficient appearance alignment with holistic context reasoning to robustly disambiguate in complex scenes.

\begin{table*}[htbp]
\centering
\caption{Retrieval performance across three progressive input configurations on CUHK-SYSU dataset, augmented using MLLM-generated descriptions.}

\label{tab:preliminary-analysis}
\setlength{\tabcolsep}{7.9pt}
\begin{tabular}{l cccc cccc cccc}
\toprule
& \multicolumn{4}{c }{\textbf{Cropped Image}} & \multicolumn{4}{c }{\textbf{Full Image}} & \multicolumn{4}{c}{\textbf{Full Image}} \\
& \multicolumn{4}{c }{\textbf{\& Appearance-only Text}} & \multicolumn{4}{c }{\textbf{\& Appearance-only Text}} & \multicolumn{4}{c}{\textbf{\& Context-enriched Text}} \\
\cmidrule(lr){2-5} \cmidrule(lr){6-9} \cmidrule(lr){10-13}
\textbf{Model} & \textbf{R@1} & \textbf{R@5} & \textbf{R@10} & \textbf{mAP} & \textbf{R@1} & \textbf{R@5} & \textbf{R@10} & \textbf{mAP} & \textbf{R@1} & \textbf{R@5} & \textbf{R@10} & \textbf{mAP} \\
\midrule

\multicolumn{13}{l}{\textit{Domain-specific TBPR models (zero-shot)}} \\
RDE~\cite{qin2024noisy} (ViT-B/16)      &40.62 & 79.19 & 84.18 & 57.34 & 
21.17 & 50.09 & 62.28 & 35.19 &
31.82 & 69.58 & 81.32 & 48.53\\
\rowcolor{gray!10}
IRRA~\cite{jiang2023cross} (ViT-B/16)     & 41.62 & 79.45 & 84.73 & 58.27 & 
23.23 & 53.47 & 66.09 & 37.76 &
35.78 & 74.68 & 85.39 & 53.04\\

\midrule
\multicolumn{13}{l}{\textit{Domain-specific TBPR models (fine-tuned)}} \\
IRRA~\cite{jiang2023cross} (ViT-B/16)     & 47.60 & 84.00 & 89.14 & 63.63 & 33.76 & 73.69 & 84.40 & 51.37 & 66.61 & 98.53 & 99.60 & 80.72 \\
RDE~\cite{qin2024noisy} (ViT-B/16)      & 48.73 & 83.78 & 88.88 & 63.93 & 35.12 & 73.47 & 82.97 & 51.98 & 67.23 & 98.31 & 99.60 & 81.05 \\
\rowcolor{gray!10}
IRRA~\cite{jiang2023cross} (ViT-L/14)     & 50.57 & 85.54 & 89.36 & 66.01 & 36.81 & 76.66 & 86.13 & 54.63 & 71.41 & 98.86 & 99.93 & 83.56 \\
\midrule
\multicolumn{13}{l}{\textit{Multimodal large model (zero-shot)}} \\
\rowcolor{gray!10}
InternVL-8B~\cite{chen2024expanding}         & 23.38 & 40.92 & 50.79 & 31.03 & 46.32 & 80.37 & 92.13 & 59.39 & 85.27 & 97.53 & 98.90 & 90.63 \\
\bottomrule
\end{tabular}
\end{table*}

\section{Preliminary Analysis of Current Models on Text-Based Person Retrieval} \label{sec:Preliminary Analysis}
Although text-based person retrieval has made significant strides driven by vision-language pretraining models, existing methods~\cite{jiang2023cross,bai2023rasa,cao2024empirical,ergasti2024mars,qin2024noisy} predominantly rely on appearance descriptions and cropped pedestrian images, overlooking the potential value of scene context. 
To empirically substantiate this oversight and demonstrate the complementary strengths of domain-specific TBPR models and MLLMs in handling increasing scene complexity, which thereby motivating our two-stage SA-Person framework, we conduct a preliminary analysis on the augmented CUHK-SYSU~\cite{xiao2017joint}.

\subsection{Analysis Configuration}





Originally featuring full-scene images with annotated target pedestrian bounding boxes, we augment CUHK-SYSU with MLLM-generated textual descriptions, with two types per target: (1) appearance-only descriptions, capturing visual attributes such as clothing, accessories, and pose; and (2) context-enriched descriptions, augmenting these with contextual details that include surroundings, spatial relations and interactions. 
This enables a gradual incorporation of scene context, tracing the shift from appearance-focused to context-aided retrieval
through three progressive input configurations:


\begin{enumerate}[label=\arabic*)]
    \item Appearance-only text \& cropped image: Relies solely on appearance-focused descriptions and the target's cropped bounding box region, isolating appearance alignment.
    \item Appearance-only text \& full image: Pairs the same descriptions with complete scene images, introducing scene information without explicit contextual guidance.
    \item Context-enriched text \& full image: Combines augmented descriptions with full scenes, evaluating joint reasoning over appearance and environmental semantics.
\end{enumerate}


Evaluated models span two architectural paradigms: (1) state-of-the-art domain-specific TBPR models, including IRRA~\cite{jiang2023cross} and RDE~\cite{qin2024noisy}, which use dual-encoder frameworks. For this analysis, we finetune these models on the augmented CUHK-SYSU dataset, adapting to each configuration's input modality (e.g., holistic visuals for the last setting). We also evaluate their zero-shot performance using pre-trained weights from CUHK-PEDES. (2) Multimodal large language models (MLLMs), such as InternVL-8B~\cite{chen2024expanding}, which support zero-shot retrieval without modality-specific adaptation. 
However, current MLLMs struggle with gallery-wide retrieval across vast image galleries, constrained by their bounded context capacities that restrict simultaneous handling of extensive images~\cite{niu2025chatreid}.
To address this, we transform the complex retrieval task into a multi-round scoring and ranking process through engineering enhancements. 
\subsection{Performance Analysis}



Table~\ref{tab:preliminary-analysis} presents results across the three progressive configurations, revealing critical insights into the role of scene context in TBPR and highlighting the complementary strengths of domain-specific TBPR models and MLLMs when handling scene complexity. 

In the appearance-only text \& cropped image setting, domain-specific models dominate.
Zero-shot IRRA pre-trained on CUHK-PEDES achieves 41.62\% R@1, rising to 50.57\% after fine-tuning, both surpassing zero-shot InternVL-8B's 23.38\% R@1. 
 Their specialized architecture and extensive appearance-focused pretraining yield highly discriminative embeddings for clean, isolated regions. In contrast, zero-shot MLLMs lack domain alignment, producing coarser cross-modal representations.

However, switching to appearance-only text \& full image setting, full scenes without contextual support, domain-specific models exposes a sharp vulnerability.
IRRA with ViT-L/14 drops from 50.57\% to 36.81\% R@1, and zero-shot domain models fall to $\sim$23\%.
Distracting environmental elements like cluttered backgrounds and occlusions overwhelm appearance-based embeddings lacking semantic safeguards. In contrast, zero-shot InternVL-8B rises to 46.32\% R@1, leveraging inherent scene understanding to mitigate interference. 

Shifting to the context-enriched text \& full image setting further underscores the value of scene context, where contextual augmentation markedly enhances retrieval accuracy across models. 
As shown in Figure~\ref{fig:retrieval_comparison_preliminary}, the contextual support enables refined disambiguation.
Zero-shot IRRA yields only 35.78\% R@1 due to limited generalization to context-enriched descriptions. 
Fine-tuning boosts R@1 to 71.41\%, a 99\% relative gain, yet still falls short of the zero-shot MLLM baseline, where InternVL-8B attains 85.27\% R@1, demonstrating clear superior inherent generalization and seamless integration of local appearance with global elements like landmarks and relational cues. Domain-specific models, despite their improvements, cannot fully leverage these aspects.

\subsection{Time Complexity Analysis}
For computational cost, MLLMs incur substantially higher expenses than domain-specific models due to the multi-round gallery retrieval process. 
Let $\mu_s$ and $\mu_m$ denote the per-image time units for domain-specific models and MLLMs, respectively, with $\mu_m \approx 2$ orders of magnitude larger than $\mu_s$ due to intensive multimodal processing. 
Domain-specific models require $O(N \mu_s)$ for a gallery of size $N$.
MLLMs demand an initial $O(N)$ traversal for match scores at $O(N \mu_m)$ time, plus $O(M/B)$ batch ranking operations with per-batch time $T_b = \mu_m \times B$, where $M$ is the candidate set size and $B$ is the batch size. 
The total time thus simplifies to $O(N \mu_m + K \mu_m)$. This yields an overall per-query time orders of magnitude higher than that of domain-specific models, scaling linearly with $N$ and limiting scalability for large galleries.

In summary, this analysis reveals a key contrast: domain-specific models excel in straightforward appearance tasks but demonstrate poor zero-shot generalization and struggle with scene variations, while MLLMs deliver superior and robust holistic reasoning in context-rich environments at a higher computational cost. The widening performance gaps as scene context increases underscore the value of approaches that connect detailed region matching and broader reasoning, directly motivating our development of SCENEPERSON-13W and the two-stage SA-Person framework to harness these complementary strengths for robust, real-world TBPR.


\begin{figure*}[t]
    \centering
    \includegraphics[width=1\linewidth]{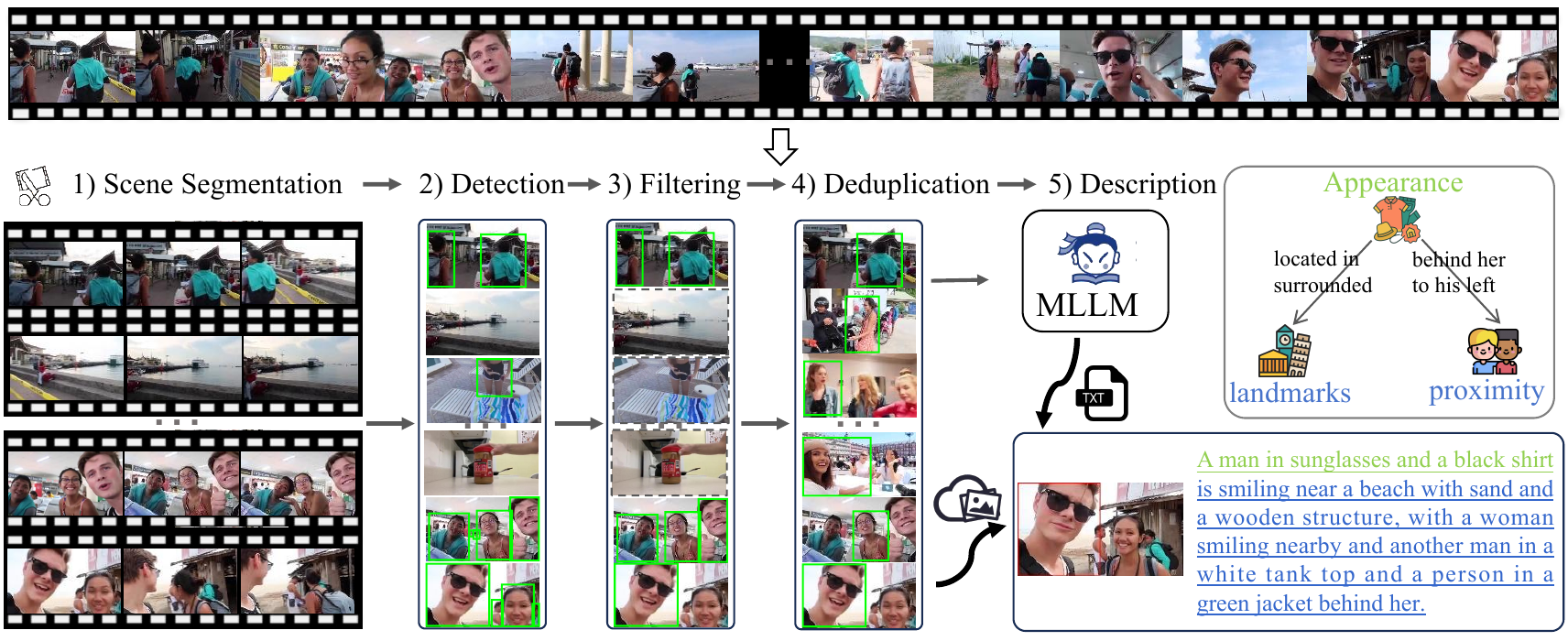}
    \caption{Overview of the {\our} construction pipeline. The construction pipeline involves scene segmentation, pedestrian detection and tracking, completeness filtering, image deduplication, and description generation. For each retained pedestrian, a description is generated based on the full image with a highlighted target, capturing their appearance, spatial location, and relationships with surrounding elements. }
    \label{fig:data construction}
\end{figure*}
\begin{table*}[t]
\centering
\caption{Statistics of person retrieval datasets. SCENEPERSON-13W shows a substantial advancement over existing person retrieval datasets by combining fine-grained appearance details with rich global scene contexts.}
\label{tab:dataset_comparison}
\setlength{\tabcolsep}{20pt} 
\resizebox{1\textwidth}{!}{
\begin{tabular}{l c c c c}
\toprule
\textbf{Dataset} & \textbf{Query Modality} & \textbf{Gallery Modality} & \textbf{\#Description} & \textbf{\#Image}  \\
\midrule
{Market-1501}~\cite{zheng2015scalable} & Image (Cropped) & Image (Cropped) & -- & 32,668  \\
{CUHK03}~\cite{li2014deepreid} & Image (Cropped) & Image (Cropped) & -- & 14,097   \\
{CUHK-SYSU}~\cite{xiao2017joint} & Image (Cropped) & Image (Full) & -- & 18,184  \\
{PRW}~\cite{zheng2017person} & Image (Cropped) & Image (Full) & -- & 11,816   \\
\hdashline 
{CUHK-PEDES}~\cite{li2017person} & Text (Appearance) & Image (Cropped) & 80,412 & 40,206  \\
{ICFG-PEDES}~\cite{ding2021semantically} & Text (Appearance) & Image (Cropped) & 54,522 & 54,522  \\
{CUHK-SYSU-TBPS}~\cite{zhang2023text} & Text (Appearance) & Image (Full) & 35,960 & 18,184   \\
{PRW-TBPS}~\cite{zhang2023text} & Text (Appearance) & Image (Full) & 19,009 & 11,246 \\
\rowcolor{gray!15}
\textbf{\our} & Text (\textbf{Appearance+scene}) & Image (Full) & 131,438 & 131,438  \\
\bottomrule
\end{tabular}
}
\end{table*}

\section{{\our} Dataset} \label{sec:dataset}

The {\our} dataset is proposed to support scene-aware text-based  person retrieval by incorporating rich scene context alongside detailed pedestrian appearance in textual descriptions.
It contains $131,438$ pedestrian instances across $112,872$ diverse scenes, with each instance accompanied by a textual description, a full-scene image, and a bounding box highlighting the pedestrian's location.
The dataset consists of $121,438$ training images and $10,000$ test images, which preserves scene complexity and pedestrian distributions with reduced redundancy, enabling comprehensive evaluation across diverse scenarios.

The construction process of the dataset illustrated in Figure~\ref{fig:data construction}, 
involves five key steps: scene segmentation, pedestrian detection and tracking, completeness filtering, deduplication in image collection, and textual description generation.

 {\bfseries Image Collection.} 
The image data in {\our} is constructed from the QVHighlights video dataset~\cite{lei2107qvhighlights} through a multi-stage processing pipeline designed to ensure data quality and diversity. 
1) Scene segmentation. Videos are segmented into distinct scenes, each representing a unique environmental context. This segmentation enables the dataset to capture a wide range of pedestrian-environment combinations, supporting comprehensive scene-aware analysis.
2) Pedestrian detection and tracking. 
Each segmented scene is then processed to detect and track pedestrians.
For each detected pedestrian instance, the frame with the highest confidence score is selected,  and the corresponding full scene image with cropped pedestrian region are retained.
3) Completeness filtering.
To enhance data quality, incomplete or noisy detections--such as those capturing only partial body parts--are filtered out using MediaPipe keypoint detection~\cite{mediapipe_pose}. Only images containing a head and at least one shoulder are preserved, ensuring the availability of essential visual features. 
4) Deduplication. Finally, a deduplication process is applied to reduce redundancy. Using CLIP~\cite{radford2021learning}, visual similarity between images is assessed. Instances that are highly similar in both the cropped pedestrian region and the full scene are discarded, promoting the uniqueness and diversity of the resulting dataset. Finally, we collect $131,438$ pedestrian instance, accompanied with $131,438$ corresponding scene images.

\begin{figure}[h]
    \centering
  \includegraphics[width=1.0\linewidth]{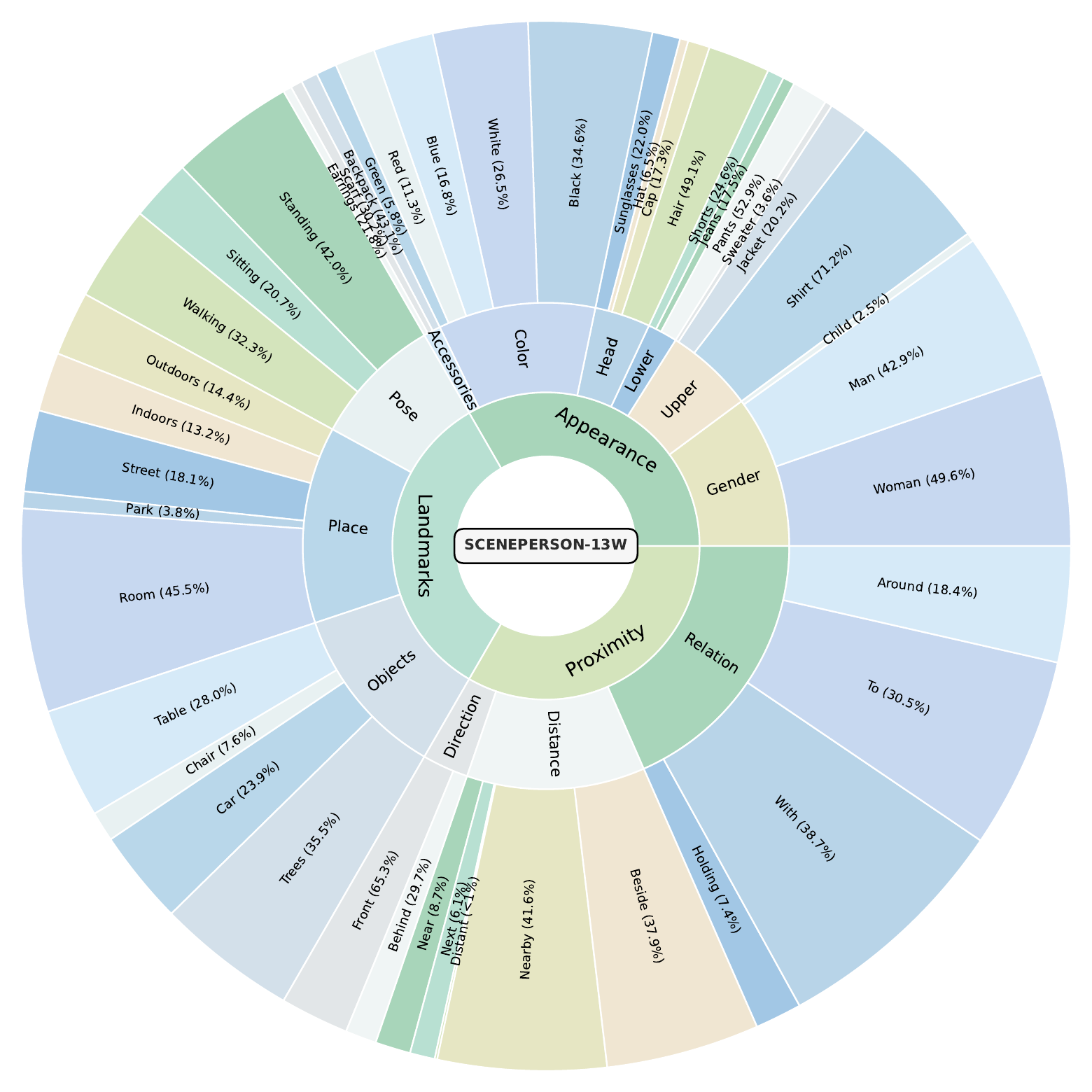}
  \caption{
  Sunburst visualization of {\our}, showing hierarchical feature distribution across Appearance, Landmarks, and Proximity. 
  }
  \label{fig:dataset_sunburst}   
\end{figure}

{\bfseries Description Generation.}
For each of the $131,438$ retained pedestrian instances, a corresponding textual description is generated using InternVL2.5-8B~\cite{chen2024expanding}.
Specifically, the model is prompted with the full-scene image in which the target pedestrian is highlighted using a red bounding box. The prompt is structured as follows:

\begin{tcolorbox}
\it
Generate a description of the individual within the red bounding box and the connection with the surroundings. Ensure that the description is a natural, fluent paragraph (under $50$ words) that a real witness would actually say to police or friends when trying to help find this exact person. Do not include text details or labels, and strictly avoid any punctuation other than commas and periods.
\end{tcolorbox}

The prompt is designed to elicit detailed descriptions by guiding the model to focus on three key aspects: the individual’s appearance, location within the scene, and relationships with surrounding elements, following a structured template. 



Compared to existing person retrieval datasets\cite{zheng2015scalable,li2014deepreid,xiao2017joint,zheng2017person,li2017person,ding2021semantically,zhang2023text}, {\our} offers a significant advancement by integrating fine-grained appearance details with rich global scene context across diverse scenarios and surpassing prior datasets in scale, as shown in Table ~\ref{tab:dataset_comparison}. 
And figure~\ref{fig:dataset_sunburst} further illustrates the hierarchical feature distribution in {\our}, enabling precise, context-rich descriptions that jointly capture fine-grained appearance and complex scene context in real-world settings.



\begin{figure*}[t]
    \centering
    \includegraphics[width=1\linewidth]
    {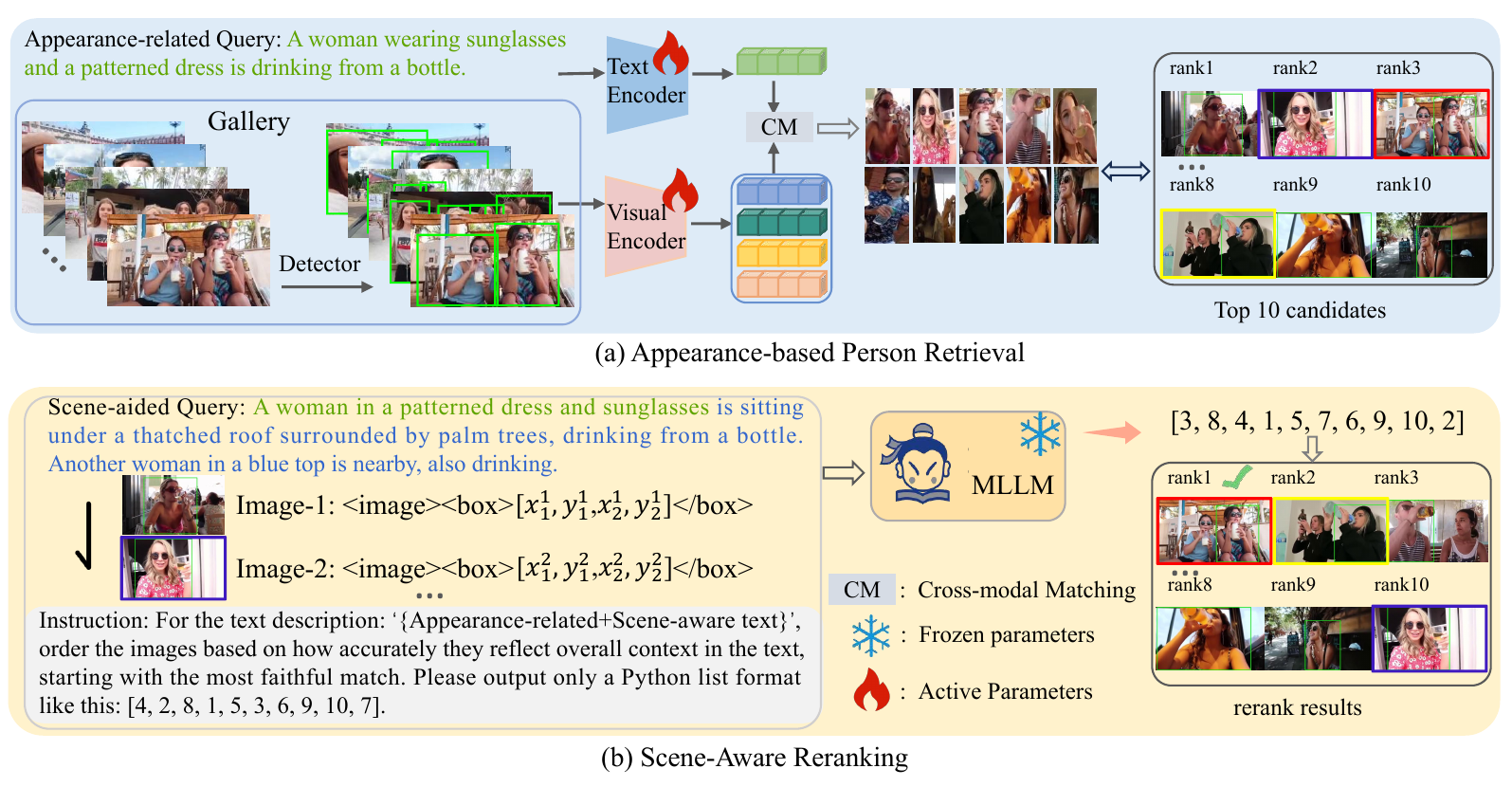}
    \caption{ 
    Overview of the proposed SA-Person framework.
     The first stage, appearance-based person retrieval, aligns pedestrian-specific regions with appearance-related descriptions for initial retrieval. The second stage, scene-aware reranking, re-ranks the top-$K$ candidates using scene-aided text and the full-scene image.
    }
    \label{fig:framework}
\end{figure*}

\section{Method} \label{sec:method}
Motivated by the insights revealed in Section~\ref{sec:Preliminary Analysis}, 
we propose SA-Person, a straightforward yet effective two-stage framework, as shown in Figure~\ref{fig:framework}. 
The first stage aligns pedestrian-specific regions with appearance-related textual descriptions to retrieve a compact set of candidates. The second stage refines the top candidates by incorporating global scene context via a training-free scene-aware re-ranking method.

\subsection{Appearance-based Retrieval}


{\bfseries Appearance Embedding Training.}
The first stage focuses on retrieving a shortlist of candidate individuals from the gallery set based solely on their visual appearance, given the appearance-related content of the textual query.
For a gallery of full-scene images \(\mathcal{G} = \{I_1, I_2, \dots, I_N\}\) and a text $T$ describing the target person, 
we use appearance-related text that describes the target person’s visual traits, 
such as clothing, hairstyle, or accessories (e.g., \textit{A man wearing a red jacket and blue jeans is laughing}), denoted as \({T}_{\text{app}}\). This ensures that the retrieval process prioritizes appearance-focused matching, excluding contextual or environmental cues at this stage.

To process the visual input \(\mathcal{G}\), we employ an off-the-shelf pedestrian detector to identify and crop all individuals within each scene image \(I_i\). This results in a set of bounding boxes \(\mathcal{B}_i = \{B_{i}^1, B_{i}^2, \dots, B_{i}^{m_i}\}\), where \(B_{i}^j = (x_j, y_j, w_j, h_j)\) represents the coordinates of the \(j\)-th detected person, ${m_i}$ represents the number of person detected in \(I_i\). Each cropped region \(B_{i}^j\) is treated as an independent person image for feature extraction.



We employ a dual-encoder architecture, initialized with a pre-trained vision-language model,
to encode the cropped person images and appearance-related
textual descriptions. 
For each cropped person image $b_{ij}$ and its corresponding appearance-related textual description \({T}_{\text{app}}\), the visual and textual feature is computed as:
\begin{equation}
\label{equation1}
    f_{ij}^v = \text{Enc}_{\text{v}}(b_{ij}),
    \quad f^t = \text{Enc}_{\text{t}}({T}_{\text{app}}),
\end{equation}
where $\text{Enc}_{\text{v}}$ and $\text{Enc}_{\text{t}}$ represent visual encoder and textual encoder, respectively, \(f_{ij}^v \in \mathbb{R}^d\) represents the visual feature representation of the \(j\)-th person in image \(I_i\) and \(f^t \in \mathbb{R}^d\) is the textual feature representation aligned with the visual feature space.

To enhance cross-modal alignment, we fine-tune the dual-encoder on these appearance-related image-text pairs from \our, following the implicit relation reasoning proposed in~\cite{jiang2023cross}. 
The loss is formally expressed as:
\begin{equation}
\begin{split}
\mathcal{L}_{\text{app}} = \mathbb{E}_{(b_{ij}, \mathcal{T}_{\text{app}}, y_{ij}) \sim \mathcal{D}_{\text{app}}} \big[
    &\ell_{\text{IRR}}(f_{ij}^v, f^t, y_{ij}; \Theta) \\
  + &\ell_{\text{SDM}}(f_{ij}^v, f^t, y_{ij}; \Theta) \\
  + &\ell_{\text{ID}}(f_{ij}^v, f^t, y_{ij}; \Theta) \big],
\end{split}
\end{equation}
where \(\mathcal{D}_{\text{app}}\) is the set of appearance-related image-text pairs with identity labels \(y_{ij}\), \(\ell_{\text{IRR}}\), \(\ell_{\text{SDM}}\), and \(\ell_{\text{ID}}\) 
are the respective loss components as defined in~\cite{jiang2023cross}, and \(\Theta\) denotes the parameters of the image and text encoders. The expectation is computed over the training set, ensuring discriminative and robust feature representations.


{\bfseries Retrieval.}
During inference, we first apply the pedestrian detector to a gallery of full-scene images $\mathcal{G} = \{\mathbf{I}_j\}_{j=1}^M$, extracting all detected person crops across the gallery. This yields a set of $K$ cropped person images $\mathcal{C} = \{\mathbf{C}_k\}_{k=1}^K$, where each $\mathbf{C}_k$ originates from a full-scene image $\mathbf{I}_j \in \mathcal{G}$. For each crop $\mathbf{C}_k$, we also retain its bounding box $\mathbf{B}_k$ and its source image identifier $j = s(k)$, indicating that $\mathbf{C}_k$ was cropped from $\mathbf{I}_{s(k)}$.

Each cropped image \(C_k\) is encoded into \(f_k^v\) using the fine-tuned image encoder and the appearance-related text \(\mathcal{T}_{\text{app}}\) is encoded into \(f^t\), as defined previously in Eq.\eqref{equation1}.
The similarity score between the textual feature and each person image is computed as:
\begin{equation}
S_k = s(\mathbf{f}^t, \mathbf{f}_k^v) = \frac{\mathbf{{f}^t}^\top \mathbf{f}_k^v}{\|\mathbf{f}^t\| \|\mathbf{f}_k^v\|}.
\end{equation}
The top $K$ candidates with the highest similarity scores are selected:
\begin{equation}
\mathcal{C}_{\text{top}} = \left\{ (\mathbf{C}_k, \mathbf{B}_k, \mathbf{I}_{s(k)}) \mid k \in \text{argmax}_{K}   S_k \right\}.
\end{equation}
$\mathcal{C}_{\text{top}}$ will be passed to the subquent re-ranking stage for further improvement based on scene understanding.

\subsection{Scene-aware Re-ranking}

While the appearance-based retrieval stage provides a strong initial filter by matching at the region-level focused on appearance, it can be limited in fully leveraging broader contextual cues compared to MLLMs, which offer superior holistic reasoning in full-scene settings.
To leverage these complementary strengths without the overhead of gallery-wide processing, we introduce SceneRanker, a scene-aware ranking module that applies a MLLM to refine the top candidates from the appearance-based retrieval stage, without any task-specific fine-tuning.
This confines expensive inference to a compact subset for scalable disambiguation.


{\bfseries Candidate-to-Scene Mapping.} 
After obtaining a ranked list of cropped person candidates from the coarse retrieval stage, we map each crop back to its corresponding full-scene image. Specifically, we link each candidate crop
$\mathbf{C}_m$ to the source image $\mathbf{I}_{s(m)} \in \mathcal{G}$  it was extracted from, along with its corresponding bounding box by the source index $s(m)$.
This mapping step bridges the pedestrian-specific region and its broader scene context.

The associated bounding box $\mathbf{B}_m$ specifies the location of the person within $\mathbf{I}_m$. It tends to  
serve as a visual prompt 
that anchors the model’s attention on each candidate individual while enabling contextual reasoning about their surrounding scene, such as spatial configurations, co-occurring objects, and relational cues. This scene-aware grounding allows the model to evaluate not only the appearance of the person but also how well their broader context aligns with the textual description.

{\bfseries Scene-aware Multimodal Prompt.}
To guide SceneRanker to perform reranking, we construct a unified prompt that integrates both the annotated visual inputs and an instructional text.
The MLLM input is a structured multimodal prompt combining the $K$ full-scene images \(\{I_m\}_{m=1}^{K}\), their corresponding bounding box annotations \(\{\mathbf{B}_m\}_{m=1}^{K}\), and the complete text $T$. The visual prompt is formatted as a markdown sequence, where each full-scene image is paired with its bounding box annotation:
\begin{equation*} 
\mathbf{P}_v = \text{Concat}\left(\{\text{Image-}m: \langle \text{image} \rangle \langle \text{box} \rangle \mathbf{B}_m \langle / \text{box} \rangle\}_{m=1}^{K}\right),
\end{equation*} 
where $\langle \text{image} \rangle$ is a placeholder for $\mathbf{I}_m$, and $\langle \text{box} \rangle \mathbf{B}_m \langle / \text{box} \rangle$ denotes the spatial region of interest corresponding to the candidate.
% ...
And an instructional prompt is appended to define the ranking task.
Specifically, the final multimodal prompt  $\mathbf{P}$  takes the following form:



% ...



\begin{tcolorbox}
\textbf{Image-1:} \texttt{<image>} \texttt{<box>}$B_1$\texttt{</box>}

\textbf{Image-2:} \texttt{<image>} \texttt{<box>}$B_2$\texttt{</box>}

\textbf{Image-3:} \texttt{<image>} \texttt{<box>}$B_3$\texttt{</box>}

...

\textbf{Image-K:} \texttt{<image>} \texttt{<box>}$B_K$\texttt{</box>}

\vspace{0.5em}

\textbf{Instruction:} For the text description: ``$T$'', order the images based on how accurately they reflect the overall context in the text, starting with the most faithful match. 
Please output only a Python list format like this: 
[4, 2, 8, 1, 5, 3, 6, 9, 10, 7, $\cdots$, $K$].
\end{tcolorbox}





{\bfseries Re-ranking.}
The SceneRanker processes the multimodal input $(\{\mathbf{I}_m, \mathbf{B}_m\}_{m=1}^{K}, \mathbf{P})$ to produce a ranked list of indices:
\begin{equation*} 
\mathbf{R} = \text{SceneRanker}(\{\mathbf{I}_m, \mathbf{B}_m\}_{m=1}^{K}, \mathbf{P}),
\end{equation*}
where $\mathbf{R} = [m_1, m_2, \ldots, m_{10}]$ represents the reordered indices of the $K$ candidates, with $m_1$ being the index of most semantically aligned with $T$. 

This stage enables the model to reason over contextual cues such as spatial layout, actions, and inter-person relations, 
offering complementary insights that are absent in appearance-based retrieval.
The reranking output serves as the final retrieval results of our framework, integrating both localized appearance matching and global scene understanding.

\section{Experiments} \label{sec:exp}
\subsection{Experimental Setup} \label{sec:exp_Setup}
{\bfseries Dataset and Evaluation Metrics.}
Experiments are conducted on four benchmarks: the proposed {\our} dataset, augmented CUHK-SYSU~\cite{xiao2017joint}, Urban1K~\cite{zhang2024long}, and a 5K subset of ShareGPT4V~\cite{chen2024sharegpt4v}. Unless otherwise specified, performance metrics are reported on {\our}.
Following prior retrieval works~\cite{bai2023rasa,cao2024empirical}, we evaluate overall retrieval performance using Rank@k (R@k) for k=1, 5, and 10, respectively, and mean Average Precision (mAP). R@k measures the proportion of queries where the target individual is correctly identified within the top k candidates, while mAP quantifies the ranking quality across all queries.

\begin{table*}[t]
\centering
\caption{Comparison of the proposed SA-Person with state-of-the-art methods.}
\label{tab:performance_comparison}
\setlength{\tabcolsep}{12pt} 
\resizebox{1\textwidth}{!}{
\begin{tabular}{lccccccc}
\toprule
\textbf{Method} & \textbf{Venue} & \textbf{Focus} & \textbf{Backbone} & \textbf{R@1 (\%)} & \textbf{R@5 (\%)} & \textbf{R@10 (\%)} & \textbf{mAP} \\
\midrule
\multicolumn{7}{l}{\textit{Region-level:}} \\
    RaSa~\cite{bai2023rasa} & IJCAI'23 & Appearance & ViT-B/16 & 62.51 & 83.83 & 88.09 & 72.05 \\
    IRRA~\cite{jiang2023cross} & CVPR'23 & Appearance & ViT-B/16 & 60.57 & 83.35 & 87.86 & 70.65 \\
    IRRA~\cite{jiang2023cross} & CVPR'23 & Appearance & ViT-L/14 & 63.37 & 84.54 & 88.67 & 72.71 \\
    TBPS-CLIP~\cite{cao2024empirical} & AAAI'24 & Appearance & ViT-B/16 & 50.29 & 75.24 & 81.58 & 61.54 \\
    RDE~\cite{qin2024noisy} & CVPR'24 & Appearance & ViTB/16 & 61.26 & 83.50 & 88.23 & 71.17 \\
    MARS~\cite{ergasti2024mars} & TOMM'24 & Appearance & ViTB/16 & 62.93 & 84.26 & 88.65 & 72.43 \\
    FMFA~\cite{yin2025cross} & arXiv'25 & Appearance & ViTB/16 & 59.77 & 83.22 & 88.05 & 70.09  \\
\midrule
\multicolumn{7}{l}{\textit{Global-level:}} \\
    CLIP~\cite{radford2021learning} & ICML'21 & Scene-aided & ViT-B/16 & 56.37 & 80.73 & 87.16 & 67.22 \\
    CLIP~\cite{radford2021learning} & ICML'21 & Scene-aided & ViT-L/14 & 63.51 & 85.53 & 90.28 & 73.33 \\
    LongCLIP~\cite{zhang2024long}   & ECCV'24 & Scene-aided & ViT-B/16 & 58.90 & 82.41 & 88.23 & 69.42 \\
    LongCLIP~\cite{zhang2024long}   & ECCV'24 & Scene-aided & ViT-L/14 & 65.75 & 86.22 & \textbf{90.95} & 75.05 \\
\midrule
\multicolumn{7}{l}{\textit{Region-to-Global Cascade:}} \\
    SA-Person & Ours & Scene-aided & ViT-B/16 & 77.42 & 86.61 & 87.86 & 81.96 \\
    \rowcolor{gray!10}
    SA-Person & Ours & Scene-aided & ViT-L/14 & \textbf{78.34} & \textbf{87.54} & 88.67 & \textbf{82.86} \\
\bottomrule
\end{tabular}
}
\end{table*}
{\bfseries Implementation Details.}
For the appearance-based retrieval stage, the pedestrian detector is composed of YOLOv5~\cite{wu2021application} for initial detection and MediaPipe~\cite{mediapipe_pose} for keypoint filtering, ensuring valid and reliable person instances.
The dual-encoder model is fine-tuned following the original IRRA~\cite{jiang2023cross} implementation, initialized with CLIP-ViT-B/16~\cite{radford2021learning} for images and CLIP text Transformer~\cite{radford2021learning} for text. The model is trained for $60$ epochs using the Adam optimizer with an initial learning rate of $10^{-5}$ and a cosine learning rate decay schedule. The image resolution of input crops is resized to $384 \times 128$ pixels. During inference, the size of candidate set $K$ selected for further processing is set to $10$.
For the scene-aware re-ranking stage, we employ InternVL 2.5-8B~\cite{chen2024expanding} as SceneRanker, without task-specific fine-tuning.
Inference is performed with a maximum of $128$ output tokens and deterministic decoding. All other settings follow the default Hugging Face Transformers~\cite{wolf2019huggingface} configuration.
All experiments are conducted on four RTX A40 GPUs with 48 GB of memory each.

\subsection{Comparison with State-of-the-art Methods}
We compare the proposed SA-Person with a range of state-of-the-art retrieval baselines, grouped by their retrieval strategies and visual grounding scope. Region-level methods solely focus on cropped pedestrian images, emphasizing appearance cues but lacking broader scene context. Global-level methods operate on the entire scene with full textual descriptions, capturing contextual information but often sacrificing accurate grounding of pedestrian appearance. In contrast, SA-Person adopts a region-to-global retrieval paradigm, initially matching pedestrian-specific regions to ensure discriminative grounding, followed by scene-aware reranking to incorporate semantic information from the broader environment.
As shown in Table~\ref{tab:performance_comparison}, SA-Person consistently outperforms all baselines by a large margin. 
For region-level methods, SA-Person yields a significant boost of 14.97\% at R@1 and 10.15\% at mAP over the strongest method IRRA (ViT-L/14). 
For global-level methods, SA-Person exceeds CLIP (ViT-L/14) by 14.83\% at R@1 and 9.53 at mAP, and outperforms the best-performing LongCLIP (ViT-L/14) by 12.59\% at R@1 and 7.81\% at mAP.
These results collectively highlight the superiority of our region-to-global cascade retrieval strategy, which integrates region-level discrimination with scene-level context and enhance both localization and semantic reasoning for more robust text-to-person retrieval.

\begin{table}[!t]
\caption{
Performance comparison of text-based person retrieval methods with and without the proposed SceneRanker. Red numbers indicate the absolute improvement achieved by our SceneRanker.
}
\label{tab:Combined_Effective_reranker_on_reid_modified}
\setlength{\tabcolsep}{2pt}
\rowcolors{2}{gray!10}{white}
\begin{tabular}{l llll}
\toprule
\textbf{Methods} & \textbf{R@1} & \textbf{R@5} & \textbf{R@10} & \textbf{mAP} \\
\midrule
TBPS-CLIP~\cite{cao2024empirical}           & 50.29 & 75.24 & 81.58 & 61.54 \\
\quad +SceneRanker   & $72.25_{\textcolor{red}{\scriptsize +21.96}}$ & $80.23_{\textcolor{red}{\scriptsize +4.99}}$  & $81.58_{\textcolor{red}{\scriptsize +0.00}}$  & $76.54_{\textcolor{red}{\scriptsize +15.00}}$ \\
IRRA(ViT-B/16)~\cite{jiang2023cross}             & 60.57 & 83.35 & 87.86 & 70.65 \\
\quad +SceneRanker   & $77.42_{\textcolor{red}{\scriptsize +16.85}}$ & $86.61_{\textcolor{red}{\scriptsize +3.26}}$  & $87.86_{\textcolor{red}{\scriptsize +0.00}}$  & $81.96_{\textcolor{red}{\scriptsize +11.31}}$ \\
IRRA(ViT-L/14)~\cite{jiang2023cross}             & 63.37 & 84.54 & 88.67 & 72.71 \\
\quad +SceneRanker   & $\textbf{78.34}_{\textcolor{red}{\scriptsize +14.97}}$ & $\textbf{87.54}_{\textcolor{red}{\scriptsize +3.00}}$  & $\textbf{88.67}_{\textcolor{red}{\scriptsize +0.00}}$  & $\textbf{82.86}_{\textcolor{red}{\scriptsize +10.15}}$ \\
RDE~\cite{qin2024noisy}                  & 61.26 & 83.50 & 88.23 & 71.17 \\
\quad +SceneRanker   & $77.58_{\textcolor{red}{\scriptsize +16.32}}$ & $86.84_{\textcolor{red}{\scriptsize +3.34}}$  & $88.23_{\textcolor{red}{\scriptsize +0.00}}$  & $82.18_{\textcolor{red}{\scriptsize +11.01}}$ \\
RaSa~\cite{bai2023rasa}                 & 62.51 & 83.83 & 88.09 & 72.05 \\
\quad +SceneRanker   & $77.84_{\textcolor{red}{\scriptsize +15.33}}$ & $86.83_{\textcolor{red}{\scriptsize +3.00}}$  & $88.09_{\textcolor{red}{\scriptsize +0.00}}$  & $82.27_{\textcolor{red}{\scriptsize +10.22}}$ \\
MARS~\cite{ergasti2024mars}                 & 62.93 & 84.26 & 88.65 & 72.43 \\
\quad +SceneRanker   & $78.32_{\textcolor{red}{\scriptsize +15.39}}$ & $87.39_{\textcolor{red}{\scriptsize +3.13}}$  & $88.65_{\textcolor{red}{\scriptsize +0.00}}$  & $82.77_{\textcolor{red}{\scriptsize +10.34}}$ \\
FMFA~\cite{yin2025cross}                            & 59.77 & 83.22 & 88.05 & 70.09 \\
\quad +SceneRanker                         & $77.35_{\textcolor{red}{\scriptsize +17.58}}$ & $86.59_{\textcolor{red}{\scriptsize +3.37}}$ & $88.05_{\textcolor{red}{\scriptsize +0.00}}$ & $81.95_{\textcolor{red}{\scriptsize +11.86}}$ \\
\bottomrule
\end{tabular}
\end{table}


\subsection{Compatibility of the Proposed SceneRanker}

The proposed SceneRanker, exploiting global scene context for precise disambiguation in the scene-aware reranking stage, is highly compatible and can be seamlessly adapted to various text-based person retrieval methods. Specifically, these methods can function as the first stage of appearance-based retrieval in the SA-Person framework.
We evaluate the compatibility of SceneRanker in Table~\ref{tab:Combined_Effective_reranker_on_reid_modified}, where it consistently demonstrates performance improvements. Across all baselines, SceneRanker delivers over a 14\% improvement in R@1 and more than a 10\% increase in mAP. Notably, it achieves significant gains for TBPS-CLIP, with R@1 increasing by 21.96\% (from 50.29\% to 72.25\%).
These consistent enhancements highlight SceneRanker as a flexible and effective component that integrates seamlessly with diverse text-based person retrieval methods. By leveraging complementary contextual cues beyond standalone pedestrian appearance, SceneRanker refines candidate selection and significantly improves retrieval results.

\subsection{Ablation Studies}
We conduct 
ablation studies to evaluate the effectiveness of the proposed SceneRanker, focusing on (1) the prompt design strategy, (2) the choice of MLLMs, and (3) the candidate size in the first-stage retrieval.

{\bfseries Different Prompt Design.}
We compare three strategies designed to enhance spatial grounding of pedestrian-specific region information in scene-aware multimodal prompts for person retrieval:

\begin{table}[!b]
\caption{
Performance of SceneRanker with various text-based person retrieval methods under different prompt designs. Red numbers indicate the improvement brought by each prompt design.
}

\setlength{\tabcolsep}{1pt}
\rowcolors{2}{gray!10}{white} 
\centering 
\footnotesize 
\begin{tabular}{l llll}
\toprule
\textbf{Methods} & \textbf{R@1} & \textbf{R@5} & \textbf{R@10} & \textbf{mAP} \\
\midrule

TBPS-CLIP~\cite{cao2024empirical}           & 50.29 & 75.24 & 81.58 & 61.54 \\
\quad +NP   & $71.41_{\textcolor{red}{\scriptsize +21.12}}$ & $80.19_{\textcolor{red}{\scriptsize +4.95}}$  & $81.58_{\textcolor{red}{\scriptsize +0.00}}$  & $76.03_{\textcolor{red}{\scriptsize +14.49}}$ \\
\quad +BOP  & $70.77_{\textcolor{red}{\scriptsize +20.48}}$ & $80.16_{\textcolor{red}{\scriptsize +4.92}}$  & $81.58_{\textcolor{red}{\scriptsize +0.00}}$  & $75.69_{\textcolor{red}{\scriptsize +14.15}}$ \\
\quad +BEP  & $\textbf{72.25}_{\textcolor{red}{\scriptsize +21.96}}$ & $\textbf{80.23}_{\textcolor{red}{\scriptsize +4.99}}$  & $\textbf{81.58}_{\textcolor{red}{\scriptsize +0.00}}$  & $\textbf{76.54}_{\textcolor{red}{\scriptsize +15.00}}$ \\
\midrule

IRRA(ViT-B/16)~\cite{jiang2023cross}        & 60.57 & 83.35 & 87.86 & 70.65 \\
\quad +NP   & $76.80_{\textcolor{red}{\scriptsize +16.23}}$ & $86.41_{\textcolor{red}{\scriptsize +3.06}}$  & $87.86_{\textcolor{red}{\scriptsize +0.00}}$  & $81.53_{\textcolor{red}{\scriptsize +10.88}}$ \\
\quad +BOP  & $75.78_{\textcolor{red}{\scriptsize +15.21}}$ & $86.39_{\textcolor{red}{\scriptsize +3.04}}$  & $87.86_{\textcolor{red}{\scriptsize +0.00}}$  & $80.99_{\textcolor{red}{\scriptsize +10.34}}$ \\
\quad +BEP  & $\textbf{77.42}_{\textcolor{red}{\scriptsize +16.85}}$ & $\textbf{86.61}_{\textcolor{red}{\scriptsize +3.26}}$  & $\textbf{87.86}_{\textcolor{red}{\scriptsize +0.00}}$  & $\textbf{81.96}_{\textcolor{red}{\scriptsize +11.31}}$ \\
\midrule

IRRA(ViT-L/14)~\cite{jiang2023cross}        & 63.37 & 84.54 & 88.67 & 72.71 \\
\quad +NP   & ${77.55}_{\textcolor{red}{\scriptsize +14.18}}$ & ${87.42}_{\textcolor{red}{\scriptsize +2.88}}$  & ${88.67}_{\textcolor{red}{\scriptsize +0.00}}$  & ${82.33}_{\textcolor{red}{\scriptsize +9.62}}$ \\
\quad +BOP  & ${76.69}_{\textcolor{red}{\scriptsize +13.32}}$ & ${87.29}_{\textcolor{red}{\scriptsize +2.75}}$  & ${88.67}_{\textcolor{red}{\scriptsize +0.00}}$  & ${81.85}_{\textcolor{red}{\scriptsize +9.14}}$ \\
\quad +BEP  & $\textbf{78.34}_{\textcolor{red}{\scriptsize +14.97}}$ & $\textbf{87.54}_{\textcolor{red}{\scriptsize +3.00}}$  & $\textbf{88.67}_{\textcolor{red}{\scriptsize +0.00}}$  & $\textbf{82.86}_{\textcolor{red}{\scriptsize +10.15}}$ \\
\midrule

RDE~\cite{qin2024noisy}                     & 61.26 & 83.50 & 88.23 & 71.17 \\
\quad +NP   & $77.13_{\textcolor{red}{\scriptsize +15.87}}$ & $86.83_{\textcolor{red}{\scriptsize +3.33}}$  & $88.23_{\textcolor{red}{\scriptsize +0.00}}$  & $81.87_{\textcolor{red}{\scriptsize +10.70}}$ \\
\quad +BOP  & $76.29_{\textcolor{red}{\scriptsize +15.03}}$ & $\textbf{86.88}_{\textcolor{red}{\scriptsize +3.38}}$  & $88.23_{\textcolor{red}{\scriptsize +0.00}}$  & $81.46_{\textcolor{red}{\scriptsize +10.29}}$ \\
\quad +BEP  & $\textbf{77.58}_{\textcolor{red}{\scriptsize +16.32}}$ & ${86.84}_{\textcolor{red}{\scriptsize +3.34}}$  & $\textbf{88.23}_{\textcolor{red}{\scriptsize +0.00}}$  & $\textbf{82.18}_{\textcolor{red}{\scriptsize +11.01}}$ \\
\midrule

RaSa~\cite{bai2023rasa}                     & 62.51 & 83.83 & 88.09 & 72.05 \\
\quad +NP   & $77.04_{\textcolor{red}{\scriptsize +14.53}}$ & $86.66_{\textcolor{red}{\scriptsize +2.83}}$  & $88.09_{\textcolor{red}{\scriptsize +0.00}}$  & $81.76_{\textcolor{red}{\scriptsize +9.71}}$ \\
\quad +BOP  & $76.24_{\textcolor{red}{\scriptsize +13.73}}$ & $86.74_{\textcolor{red}{\scriptsize +2.91}}$  & $88.09_{\textcolor{red}{\scriptsize +0.00}}$  & $81.29_{\textcolor{red}{\scriptsize +9.24}}$ \\
\quad +BEP  & $\textbf{77.84}_{\textcolor{red}{\scriptsize +15.33}}$ & $\textbf{86.83}_{\textcolor{red}{\scriptsize +3.00}}$  & $\textbf{88.09}_{\textcolor{red}{\scriptsize +0.00}}$  & $\textbf{82.27}_{\textcolor{red}{\scriptsize +10.22}}$ \\
\midrule

MARS~\cite{ergasti2024mars}                 & 62.93 & 84.26 & 88.65 & 72.43 \\
\quad +NP   & $77.26_{\textcolor{red}{\scriptsize +14.33}}$ & $87.30_{\textcolor{red}{\scriptsize +3.04}}$  & $88.65_{\textcolor{red}{\scriptsize +0.00}}$  & $82.10_{\textcolor{red}{\scriptsize +9.67}}$ \\
\quad +BOP  & $76.57_{\textcolor{red}{\scriptsize +13.64}}$ & $87.14_{\textcolor{red}{\scriptsize +2.88}}$  & $88.65_{\textcolor{red}{\scriptsize +0.00}}$  & $81.69_{\textcolor{red}{\scriptsize +9.26}}$ \\
\quad +BEP  & $\textbf{78.32}_{\textcolor{red}{\scriptsize +15.39}}$ & $\textbf{87.39}_{\textcolor{red}{\scriptsize +3.13}}$  & $\textbf{88.65}_{\textcolor{red}{\scriptsize +0.00}}$  & $\textbf{82.77}_{\textcolor{red}{\scriptsize +10.34}}$ \\

\midrule
FMFA~\cite{yin2025cross} 
    & 59.77 & 83.22 & 88.05 & 70.09 \\
\quad +NP  
    & $76.51_{\textcolor{red}{\scriptsize +16.74}}$ 
    & $86.58_{\textcolor{red}{\scriptsize +3.36}}$  
    & $88.05_{\textcolor{red}{\scriptsize +0.00}}$  
    & $81.47_{\textcolor{red}{\scriptsize +11.38}}$ \\
\quad +BOP 
    & $75.83_{\textcolor{red}{\scriptsize +16.06}}$ 
    & $86.48_{\textcolor{red}{\scriptsize +3.26}}$  
    & $88.05_{\textcolor{red}{\scriptsize +0.00}}$  
    & $81.09_{\textcolor{red}{\scriptsize +11.00}}$ \\
\quad +BEP 
    & $\mathbf{77.35}_{\textcolor{red}{\scriptsize +17.58}}$ 
    & $\mathbf{86.59}_{\textcolor{red}{\scriptsize +3.37}}$  
    & $\mathbf{88.05}_{\textcolor{red}{\scriptsize +0.00}}$  
    & $\mathbf{81.95}_{\textcolor{red}{\scriptsize +11.86}}$ \\
        
\bottomrule
\end{tabular}
\label{tab:visual_prompt_all}
\end{table}

1) \emph{Naive Prompt} (NP), which provides only the textual description and the full-scene image without explicit region information;
2) \emph{Box Overlay Prompt} (BOP), which visually highlights the candidate region with a red rectangle on the full-scene image;
3) \emph{Box Embedded Prompt} (BEP), which includes the bounding box coordinates of the candidate region in the multimodal prompt using \texttt{<box>} tags. 
As shown in Table~\ref{tab:visual_prompt_all},
SceneRanker significantly boosts performance for all methods, with R@1 improvements exceeding above 14.97\% with BEP, underscoring its compatibility and effectiveness in refining appearance-based retrieval. Across all methods, BEP consistently outperforms NP and BOP, achieving the highest R@1, suggesting that embedding bounding box coordinates enhances spatial grounding, enabling better alignment with textual descriptions.
Notably, BOP even play worse than NP across all baselines, with a representative drop of 1.02\% at R@1 and 0.51\% at mAP on IRRA with ViT-B/16, indicating that direct visual highlighting may introduce visual distractions that disrupt holistic scene understanding.

{\bfseries Different Multimodal Large Language Models.}
We evaluate the impact of integrating different 8B-scale MLLMs into the proposed SceneRanker, using the unified naive prompt to mitigate varying capabilities in processing structured inputs. As shown in Table~\ref{tab:MLLMs}, all three models, Qwen-VL 2.5, MiniCPM-V, and InternVL 2.5, consistently improve over the IRRA baseline, with respective R@1 gains of +11.7\%, +7.96\%, and +14.18\%.  
In addition, we observe a strong correlation between model capacity and performance. Taking InternVL2.5 as an example, scaling from 4B to 8B brings further improvement, with R@1 increasing from 75.21\% to 77.55\%. In contrast, smaller models below 4B such as InternVL 1B and 2B show significantly lower performance, indicating insufficient multimodal understanding. 
These results confirm SceneRanker’s effectiveness and generalizability across MLLM architectures with sufficient vision-language reasoning capacity.

\begin{table}[t]
    \caption{
    Impact of integrating different MLLMs into SceneRanker.
    Due to varying MLLM capabilities in processing structured prompts, we adopt Naive Prompt across all models to ensure fair comparison.}
    \centering
    \begin{tabular}{lccccc}
        \toprule
        \multirow{1}{*}{\textbf{Model}} & \multirow{1}{*}{\textbf{Parameters}} 
         & \textbf{R@1} & \textbf{R@5} & \textbf{R@10} & \textbf{mAP} \\
        \midrule
        IRRA (ViT-L/14) & 590 Million & 63.37 & 84.54 & 88.67 & 72.71 \\
        \midrule
       

        \rowcolor{white} 
        \multirow{2}{*}{QwenVL2.5~\cite{bai2025qwen2}} 
            & 3 Billion  & 54.57 & 78.23 & 88.67 & 65.04 \\ 
        \rowcolor{gray!10}
        \multirow{-2}{*}{QwenVL2.5~\cite{bai2025qwen2}} 
            & 8 Billion  & 75.07 & 86.34 & 88.67 & 80.45 \\
        \cmidrule(lr){1-6}
        \rowcolor{gray!10}
        MiniCPM-V-2\_6~\cite{yao2024minicpm}
        
        & 8 Billion & 71.33 & 84.66 & 88.67 & 77.78 \\ 
        \cmidrule(lr){1-6}
         \multirow{4}{*}{InternVL2.5~\cite{chen2024expanding}}
            & 1 Billion & 3.01 & 80.14 & 88.67 & 26.71 \\
            & 2 Billion & 44.14 & 75.32 & 88.67 & 56.74 \\
            & 4 Billion & 75.21 & 86.97 & 88.67 & 80.78 \\
             \rowcolor{gray!10}
            & 8 Billion & \textbf{77.55} & \textbf{87.42} & \textbf{88.67} & \textbf{82.33} \\ 
        \bottomrule
    \end{tabular}
    \label{tab:MLLMs}
\end{table}

\begin{figure}[h]
    \centering
  \includegraphics[width=0.9\linewidth]{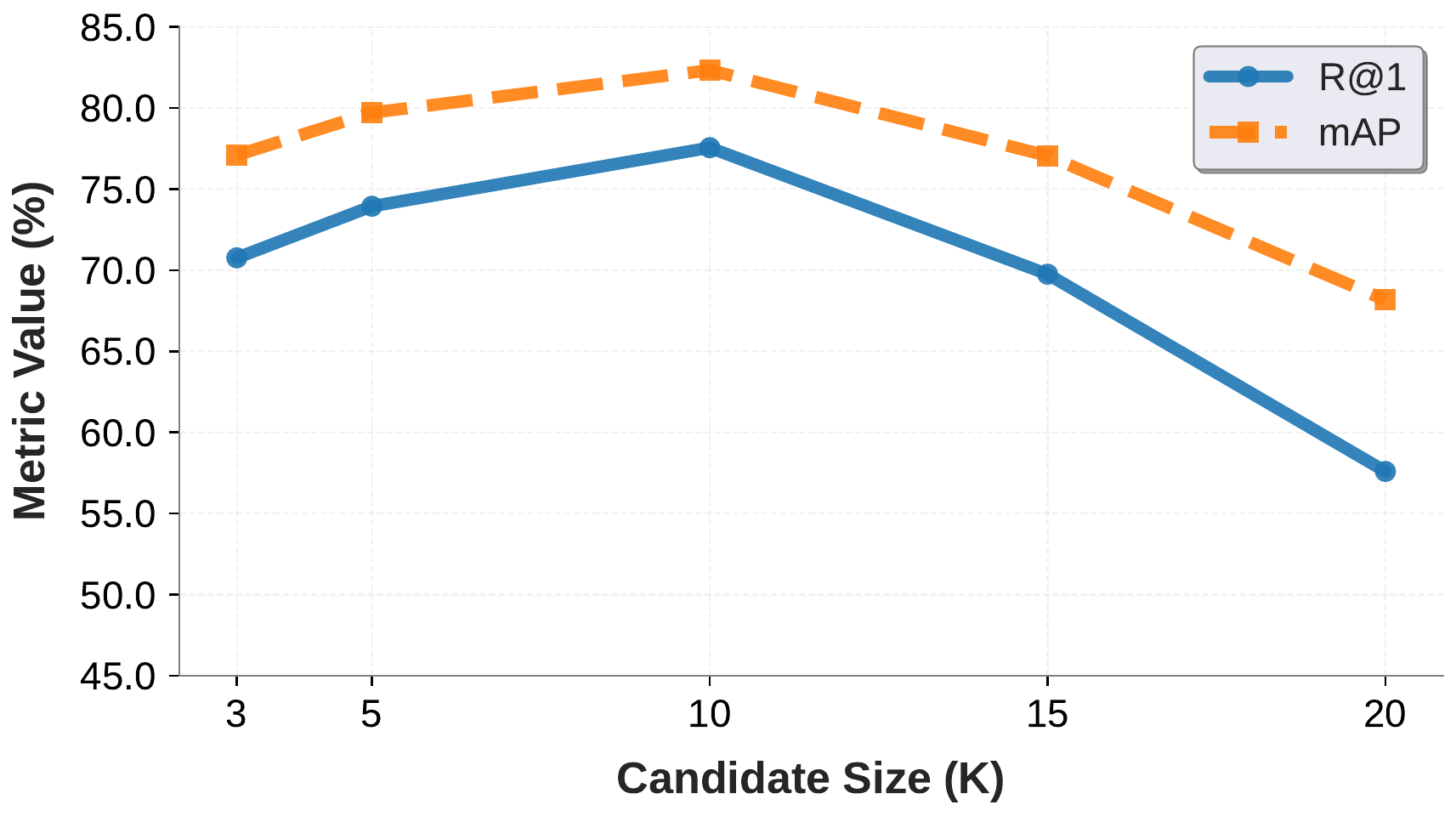}
  \caption{Impact of candidate size.}
  \label{fig:Candidate_Size}   
\end{figure}
{\bfseries Different Candidate Sizes.}
In the proposed two-stage framework SA-Person, the candidate set produced by the appearance-based retrieval stage serves as the input for the SceneRanker to perform scene-aware reranking.
The size of this set ($K$) governs the trade-off between ensuring the inclusion of the ground-truth match and minimizing the introduction of noisy distractors.
To better understand this trade-off and determine an optimal value for ($K$), we conduct a parameter analysis to systematically investigate the impact of different candidate set sizes on the final retrieval performance.
As shown in Figure ~\ref{fig:Candidate_Size}, the performance of SceneRanker exhibits a trend of first increasing and then decreasing as the candidate size changes. The best performance is achieved when the candidate size is set to 10.
Smaller candidate sets are more likely to exclude the ground-truth instance, while excessively large candidate sets introduce more irrelevant or noisy candidates. Both scenarios negatively impact performance.
These results indicate that a moderate candidate pool is optimal, as it balances retaining the ground-truth instance while minimizing noise, ultimately enhancing retrieval accuracy.

\subsection{Versatility Across Diverse Retrieval Scenarios}
To demonstrate the versatility of SA-Person, we evaluate its performance across three distinct retrieval scenarios: person retrieval in surveillance scenarios using augmented CUHK-SYSU from our preliminary analysis in Section~\ref{sec:Preliminary Analysis}, person retrieval in varied real-world environments using our SCENEPERSON-13W benchmark, and general image-text retrieval in open-domain contexts using Urban1K~\cite{zhang2024long} and a 5K subset of ShareGPT4V~\cite{chen2024sharegpt4v}.


\begin{table}[htbp]
\caption{
SceneRanker Enhancements in Surveillance Scenarios.
}
\label{tab:Surveillance}
\setlength{\tabcolsep}{2pt}
\rowcolors{2}{gray!10}{white}
\begin{tabular}{l llll}
\toprule
\textbf{Methods} & \textbf{R@1} & \textbf{R@5} & \textbf{R@10} & \textbf{mAP} \\
\midrule
IRRA (ViT-B/16)~\cite{jiang2023cross}       & 47.60 & 84.00 & 89.14 & 63.63 \\
\quad +SceneRanker                        & $54.24_{\textcolor{red}{\scriptsize +6.64}}$ & $86.35_{\textcolor{red}{\scriptsize +2.35}}$ & $89.14_{\textcolor{red}{\scriptsize +0.00}}$ & $68.03_{\textcolor{red}{\scriptsize +4.40}}$ \\

IRRA (ViT-L/14)~\cite{jiang2023cross}       & 50.57 & 85.54 & 89.36 & 66.01 \\
\quad +SceneRanker                        & $57.17_{\textcolor{red}{\scriptsize +6.60}}$ & $87.19_{\textcolor{red}{\scriptsize +1.65}}$ & $89.36_{\textcolor{red}{\scriptsize +0.00}}$ & $70.04_{\textcolor{red}{\scriptsize +4.03}}$ \\

RDE (ViT-B/16)~\cite{qin2024noisy}         & 48.73 & 83.78 & 88.88 & 63.93 \\
\quad +SceneRanker                        & $56.51_{\textcolor{red}{\scriptsize +7.78}}$ & $86.17_{\textcolor{red}{\scriptsize +2.39}}$ & $88.88_{\textcolor{red}{\scriptsize +0.00}}$ & $69.27_{\textcolor{red}{\scriptsize +5.34}}$ \\
\bottomrule
\end{tabular}
\end{table}

\begin{table}[htbp] 
\centering
\caption{SceneRanker Enhancements in General Image-Text Retrieval Scenarios. }
\label{tab:UrbanShare}
\setlength{\tabcolsep}{2pt}
\begin{tabular}{lcccccccc}
\toprule
\multirow{2}{*}{\textbf{Model (Backbone)}} 
& \multicolumn{4}{c}{\textbf{Urban1k}} 
& \multicolumn{4}{c}{\textbf{shareGPT4V}} \\
\cmidrule(lr){2-5} \cmidrule(lr){6-9}
& R@1  & R@5 & R@10  & mAP  
& R@1  & R@5  & R@10  & mAP \\
\midrule

CLIP (ViT-B/16)          & 53.10 & 78.40 & 85.20 & 64.34 & 47.60 & 73.92 & 83.12 & 59.72 \\
\quad  + SceneRanker & 72.70 & 83.00 & 85.20 & 77.69 & 59.52 & 78.40 & 83.10 & 68.19 \\
\addlinespace
CLIP (ViT-L/14)          & 55.90 & 79.50 & 86.50 & 66.81 & 51.06 & 74.72 & 83.22 & 61.87 \\
\quad  + SceneRanker & 73.10 & 83.90 & 86.50 & 78.36 & 61.34 & 78.94 & 83.22 & 69.53 \\
\bottomrule
\end{tabular}
\end{table}

\begin{figure*}[ht]
    \centering
    \includegraphics[width=0.9\textwidth]{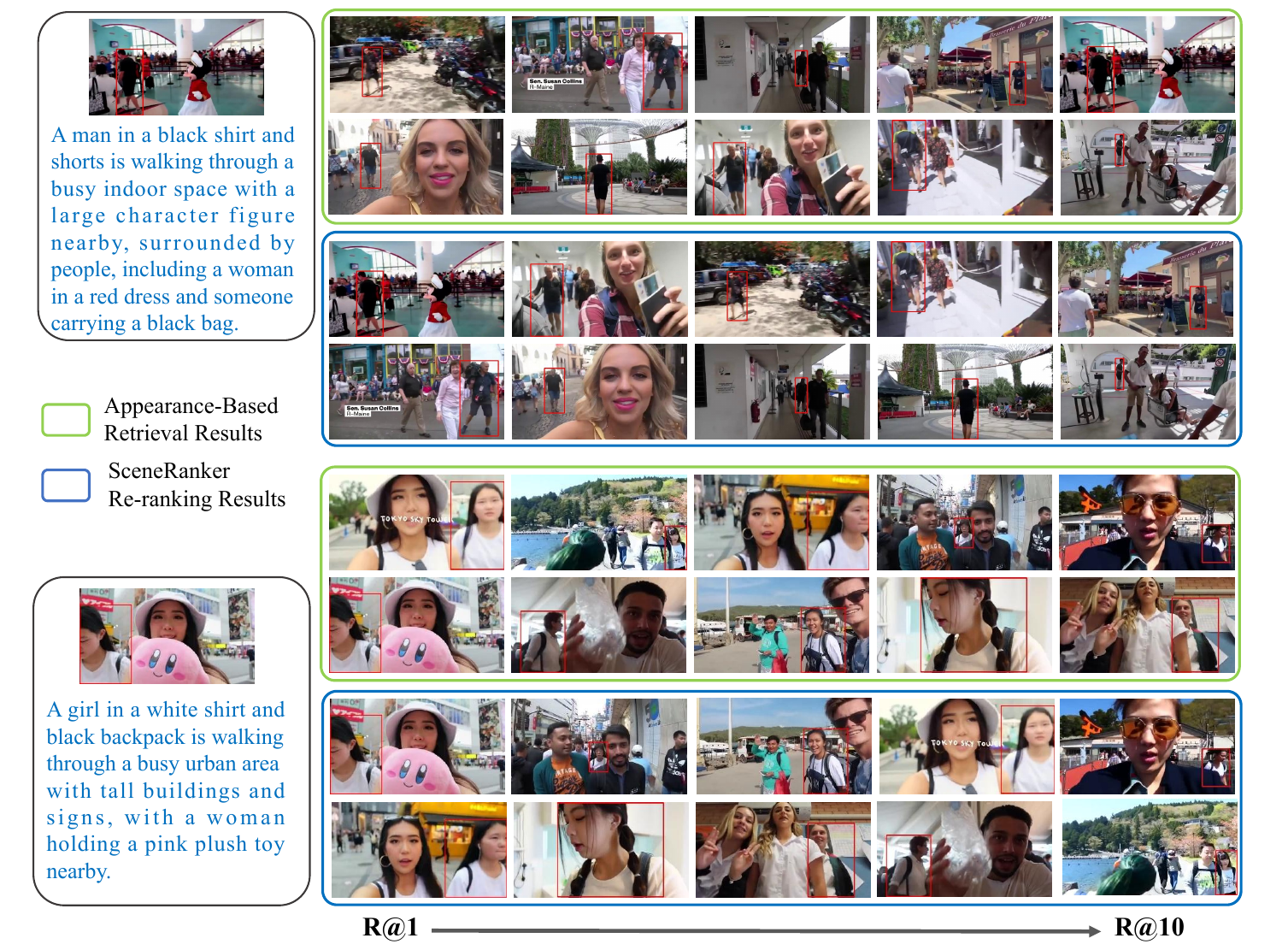} 
    \caption{Qualitative Comparison of Retrieval Results with and without SceneRanker. The leftmost column shows the text and ground-truth image. The green frame contains the retrieval results from the appearance-based stage before SceneRanker re-ranking, while the blue frame contains the retrieval results after SceneRanker re-ranking.}
    \label{fig:retrieval_comparison}
\end{figure*}


\begin{figure}[htbp]
\centering
\includegraphics[width=0.5\textwidth]{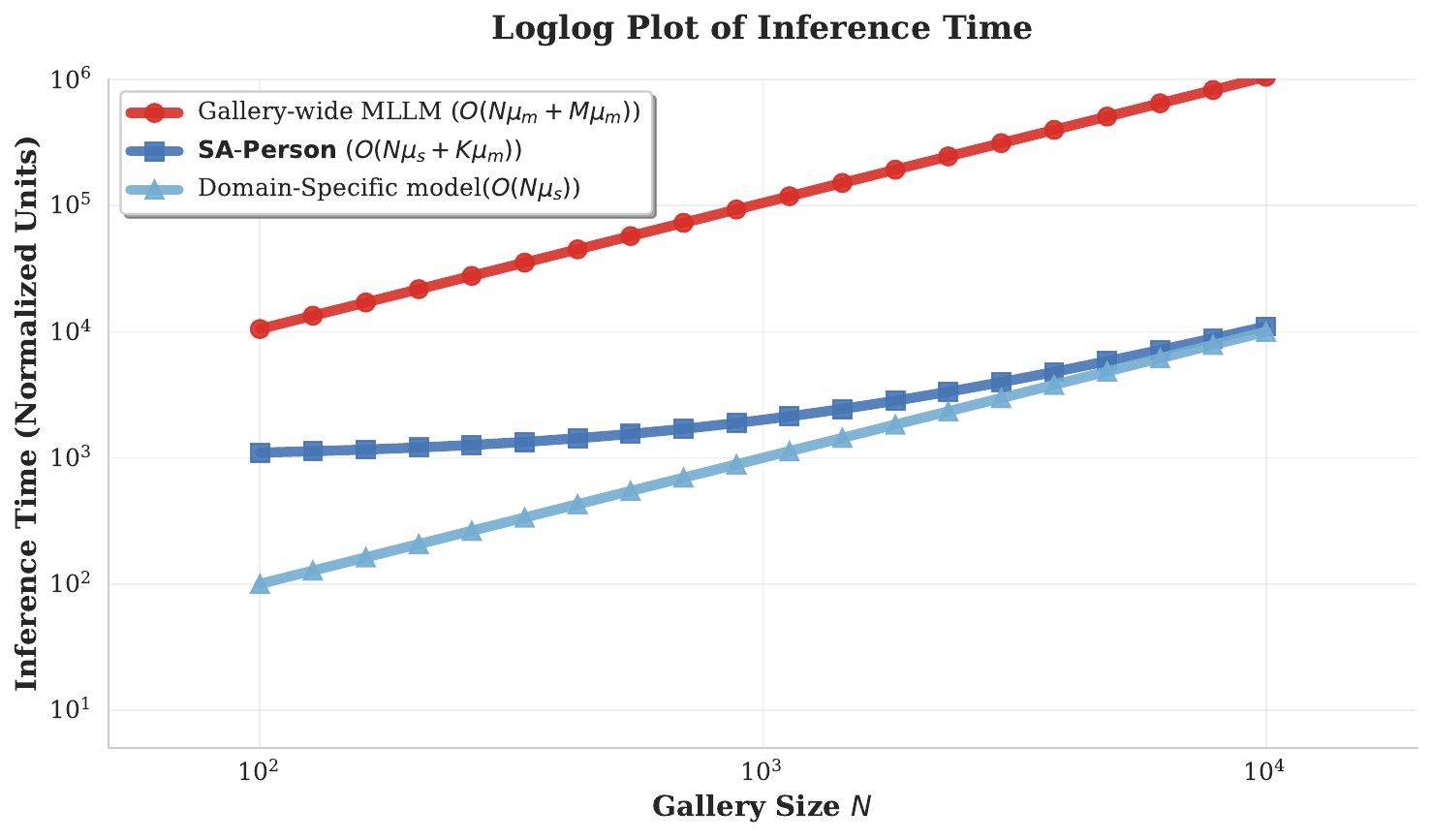}
\caption{Loglog Plot of Inference Time Across Gallery Sizes. The plot shows inference time (normalized units, log scale) versus gallery size $N$ (log scale) for three methods. }
\label{fig:time-complexity}
\end{figure}

Table~\ref{tab:Surveillance} presents SceneRanker enhancements on augmented CUHK-SYSU in surveillance scenarios. It achieves R@1 gains ranging from 6.60\% to 7.78\% and mAP improvements from 4.03\% to 5.34\% on IRRA and RDE backbones, demonstrating the module's robustness in complex surveillance environments.
On SCENEPERSON-13W, SA-Person yields large improvements over baselines as shown in Tables~\ref{tab:Combined_Effective_reranker_on_reid_modified}.
For Urban1K and ShareGPT4V, we integrate plug-and-play SceneRanker with CLIP baselines using full images and enriched descriptions. Table~\ref{tab:UrbanShare} reports retrieval performance on these datasets. SceneRanker enhances the baselines across both: R@1 improves by over 19\% and mAP by more than 12\% on Urban1K, effectively exploiting urban contextual cues such as landmarks and crowds. Comparable gains appear on ShareGPT4V, with $\sim$12\% R@1 and $\sim$8\% mAP uplifts, affirming adaptability to open-domain diversity.
These results across scenarios confirm SA-Person's broad applicability, extending beyond SCENEPERSON-13W to multimodal tasks requiring scene-aware disambiguation.

\subsection{Computational cost of inference}

Unlike prior approaches~\cite{niu2025chatreid} that rely on exhaustive MLLM-based retrieval across the entire gallery, SA-Person combines the efficiency of domain-specific models for appearance grounding with the robust contextual reasoning of MLLMs, achieving a strong balance between high retrieval performance and computational scalability.
As defined in Section~\ref{sec:Preliminary Analysis}, let $\mu_s$ and $\mu_m$ denote the per-image time units for domain-specific models and MLLMs, respectively, with $\mu_m \approx 2$ orders of magnitude larger than $\mu_s$ due to intensive multimodal processing. 
For a gallery of size $N$, Appearance-based Retrieval processes all $N$ full-scene images to yield $O(N \mu_s)$ time overall, producing a compact set of $K$ candidates ($K=10 \ll N$). Scene-aware Re-ranking then applies a single MLLM forward pass on these $K$ candidates with bounding boxes, incurring $O(K \mu_m)$ time.
The total time simplifies to $O(N \mu_s + K \mu_m)$, orders of magnitude lower than gallery-wide MLLM retrieval at $O(N \mu_m + M \mu_m)$ (where $K=10$, $M \ll N$), while surpassing pure domain-specific models at $O(N \mu_s)$ through contextual re-ranking. As illustrated in Figure~\ref{fig:time-complexity}, this design renders the MLLM stage independent of $N$, ensuring scalability for large galleries.


\subsection{Qualitative Results}

To further validate the effectiveness of the proposed SA-Person framework, we present a qualitative comparison of retrieval results before and after applying the SceneRanker module in Figure~\ref{fig:retrieval_comparison}. Without scene-aware re-ranking, the initial candidates retrieved by the appearance-based methods often include individuals with similar coarse attributes but in mismatched contexts, leading to suboptimal ranking. In contrast, SA-Person leverages an MLLM to interpret global scene semantics and re-rank these candidates accordingly.  These qualitative examples demonstrate that SA-Person effectively bridges the gap between fine-grained appearance matching and holistic scene understanding, achieving more accurate and robust text-based person retrieval.


\section{Limitations and Discussion}

While SA-Person advances scene-aware text-based person retrieval, our approach has several limitations. 
First, the textual descriptions in SCENEPERSON-13W are generated via MLLMs to enable large-scale annotation, which efficiently captures diverse appearance and contextual details but may exhibit a relatively uniform stylistic phrasing compared to the diverse variations in real-world texts.
A more effective strategy may involve incorporating diverse templates during generation to broaden stylistic coverage, and exploring directly optimizing the MLLM to select the most varied phrasing dynamically.
Additionally, SceneRanker builds on off-the-shelf MLLMs in a training-free manner for plug-and-play flexibility and broad applicability, yet task-specific fine-tuning could amplify its precision in retrieval intent alignment and scene context disambiguation. Exploring lightweight adapters or domain-adaptive pretraining would offer a promising avenue to boost performance without sacrificing zero-shot robustness.
These directions not only address potential refinements but also open avenues for extending SA-Person to interactive human-centric systems and dynamic video retrieval.

\section{Conclusion} \label{sec:conclusion}

In this paper, we tackle the challenging task of person retrieval by leveraging multimodal information from both pedestrian appearance and global scene context. Specifically, we make three key contributions. First, we present {\our}, the first large-scale dataset offering comprehensive multimodal annotations that integrate fine-grained pedestrian appearance descriptions with rich scene-level contextual information within full-scene images. 
Second, we propose SA-Person, a two-stage retrieval framework that combines discriminative appearance grounding in cropped pedestrian regions with holistic scene understanding for robust retrieval from large-scale full-scene databases. 
Extensive experimental results demonstrate that SA-Person significantly outperforms existing baselines, underscoring the critical importance of integrating both local appearance cues and global scene context for robust and effective text-based person retrieval in complex real-world scenarios.


\section*{Acknowledgments}
This work was supported in part by the InnoHK Program of the Hong Kong SAR Government; the National Natural Science Foundation of China under Grants No. 62306313, 62476188 and 62206276; the Natural Science Foundation of the Jiangsu Higher Education Institutions of China; and the Key Laboratory of New Generation Artificial Intelligence Technology \& Its Interdisciplinary Applications (Southeast University), Ministry of Education, China.



{
\bibliographystyle{IEEEtran}
\bibliography{local-bib}

\begin{thebibliography}{10}
\providecommand{\url}[1]{#1}
\csname url@samestyle\endcsname
\providecommand{\newblock}{\relax}
\providecommand{\bibinfo}[2]{#2}
\providecommand{\BIBentrySTDinterwordspacing}{\spaceskip=0pt\relax}
\providecommand{\BIBentryALTinterwordstretchfactor}{4}
\providecommand{\BIBentryALTinterwordspacing}{\spaceskip=\fontdimen2\font plus
\BIBentryALTinterwordstretchfactor\fontdimen3\font minus \fontdimen4\font\relax}
\providecommand{\BIBforeignlanguage}[2]{{%
\expandafter\ifx\csname l@#1\endcsname\relax
\typeout{** WARNING: IEEEtran.bst: No hyphenation pattern has been}%
\typeout{** loaded for the language `#1'. Using the pattern for}%
\typeout{** the default language instead.}%
\else
\language=\csname l@#1\endcsname
\fi
#2}}
\providecommand{\BIBdecl}{\relax}
\BIBdecl

\bibitem{galiyawala2021person}
H.~Galiyawala and M.~S. Raval, ``Person retrieval in surveillance using textual query: a review,'' \emph{Multimedia Tools and Applications}, vol.~80, no.~18, pp. 27\,343--27\,383, 2021.

\bibitem{li2018richly}
D.~Li, Z.~Zhang, X.~Chen, and K.~Huang, ``A richly annotated pedestrian dataset for person retrieval in real surveillance scenarios,'' \emph{IEEE transactions on image processing}, vol.~28, no.~4, pp. 1575--1590, 2018.

\bibitem{li2023dcel}
S.~Li, X.~Xu, Y.~Yang, F.~Shen, Y.~Mo, Y.~Li, and H.~T. Shen, ``Dcel: deep cross-modal evidential learning for text-based person retrieval,'' in \emph{Proceedings of the 31st ACM International Conference on Multimedia}, 2023, pp. 6292--6300.

\bibitem{zhang2016learning}
L.~Zhang, T.~Xiang, and S.~Gong, ``Learning a discriminative null space for person re-identification,'' in \emph{Proceedings of the IEEE conference on computer vision and pattern recognition}, 2016, pp. 1239--1248.

\bibitem{liu2018pose}
J.~Liu, B.~Ni, Y.~Yan, P.~Zhou, S.~Cheng, and J.~Hu, ``Pose transferrable person re-identification,'' in \emph{Proceedings of the IEEE conference on computer vision and pattern recognition}, 2018, pp. 4099--4108.

\bibitem{sarfraz2018pose}
M.~S. Sarfraz, A.~Schumann, A.~Eberle, and R.~Stiefelhagen, ``A pose-sensitive embedding for person re-identification with expanded cross neighborhood re-ranking,'' in \emph{Proceedings of the IEEE conference on computer vision and pattern recognition}, 2018, pp. 420--429.

\bibitem{xiao2017joint}
T.~Xiao, S.~Li, B.~Wang, L.~Lin, and X.~Wang, ``Joint detection and identification feature learning for person search,'' in \emph{Proceedings of the IEEE conference on computer vision and pattern recognition}, 2017, pp. 3415--3424.

\bibitem{he2018end}
Z.~He and L.~Zhang, ``End-to-end detection and re-identification integrated net for person search,'' in \emph{Asian Conference on Computer Vision}.\hskip 1em plus 0.5em minus 0.4em\relax Springer, 2018, pp. 349--364.

\bibitem{yan2021anchor}
Y.~Yan, J.~Li, J.~Qin, S.~Bai, S.~Liao, L.~Liu, F.~Zhu, and L.~Shao, ``Anchor-free person search,'' in \emph{Proceedings of the IEEE/CVF conference on computer vision and pattern recognition}, 2021, pp. 7690--7699.

\bibitem{cao2022pstr}
J.~Cao, Y.~Pang, R.~M. Anwer, H.~Cholakkal, J.~Xie, M.~Shah, and F.~S. Khan, ``Pstr: End-to-end one-step person search with transformers,'' in \emph{Proceedings of the IEEE/CVF Conference on Computer Vision and Pattern Recognition}, 2022, pp. 9458--9467.

\bibitem{yan2022exploring}
Y.~Yan, J.~Li, S.~Liao, J.~Qin, B.~Ni, K.~Lu, and X.~Yang, ``Exploring visual context for weakly supervised person search,'' in \emph{Proceedings of the AAAI Conference on Artificial Intelligence}, vol.~36, no.~3, 2022, pp. 3027--3035.

\bibitem{li2017person}
S.~Li, T.~Xiao, H.~Li, B.~Zhou, D.~Yue, and X.~Wang, ``Person search with natural language description,'' in \emph{Proceedings of the IEEE conference on computer vision and pattern recognition}, 2017, pp. 1970--1979.

\bibitem{yu2019cross}
X.~Yu, T.~Chen, Y.~Yang, M.~Mugo, and Z.~Wang, ``Cross-modal person search: A coarse-to-fine framework using bi-directional text-image matching,'' in \emph{Proceedings of the IEEE/CVF International Conference on Computer Vision Workshops}, 2019, pp. 0--0.

\bibitem{liu2023survey}
M.~Liu, Y.~Zhang, and H.~Li, ``Survey of cross-modal person re-identification from a mathematical perspective,'' \emph{Mathematics}, vol.~11, no.~3, p. 654, 2023.

\bibitem{radford2021learning}
A.~Radford, J.~W. Kim, C.~Hallacy, A.~Ramesh, G.~Goh, S.~Agarwal, G.~Sastry, A.~Askell, P.~Mishkin, J.~Clark \emph{et~al.}, ``Learning transferable visual models from natural language supervision,'' in \emph{International conference on machine learning}.\hskip 1em plus 0.5em minus 0.4em\relax PmLR, 2021, pp. 8748--8763.

\bibitem{yang2023towards}
S.~Yang, Y.~Zhou, Z.~Zheng, Y.~Wang, L.~Zhu, and Y.~Wu, ``Towards unified text-based person retrieval: A large-scale multi-attribute and language search benchmark,'' in \emph{Proceedings of the 31st ACM International Conference on Multimedia}, 2023, pp. 4492--4501.

\bibitem{li2024adaptive}
S.~Li, C.~He, X.~Xu, F.~Shen, Y.~Yang, and H.~T. Shen, ``Adaptive uncertainty-based learning for text-based person retrieval,'' in \emph{Proceedings of the AAAI Conference on Artificial Intelligence}, vol.~38, no.~4, 2024, pp. 3172--3180.

\bibitem{zuo2024ufinebench}
J.~Zuo, H.~Zhou, Y.~Nie, F.~Zhang, T.~Guo, N.~Sang, Y.~Wang, and C.~Gao, ``Ufinebench: Towards text-based person retrieval with ultra-fine granularity,'' in \emph{Proceedings of the IEEE/CVF Conference on Computer Vision and Pattern Recognition}, 2024, pp. 22\,010--22\,019.

\bibitem{luo2025graph}
B.~Luo, J.~Wang, Z.~Wang, J.~Zhu, and X.~Zhao, ``Graph-based cross-domain knowledge distillation for cross-dataset text-to-image person retrieval,'' in \emph{Proceedings of the AAAI Conference on Artificial Intelligence}, vol.~39, no.~1, 2025, pp. 568--576.

\bibitem{liu2025dm}
Y.~Liu, Z.~Liu, X.~Lan, W.~Yang, Y.~Li, and Q.~Liao, ``Dm-adapter: Domain-aware mixture-of-adapters for text-based person retrieval,'' in \emph{Proceedings of the AAAI Conference on Artificial Intelligence}, vol.~39, no.~6, 2025, pp. 5703--5711.

\bibitem{jiang2023cross}
D.~Jiang and M.~Ye, ``Cross-modal implicit relation reasoning and aligning for text-to-image person retrieval,'' in \emph{Proceedings of the IEEE/CVF Conference on Computer Vision and Pattern Recognition}, 2023, pp. 2787--2797.

\bibitem{bai2023rasa}
Y.~Bai, M.~Cao, D.~Gao, Z.~Cao, C.~Chen, Z.~Fan, L.~Nie, and M.~Zhang, ``Rasa: Relation and sensitivity aware representation learning for text-based person search,'' \emph{arXiv preprint arXiv:2305.13653}, 2023.

\bibitem{cao2024empirical}
M.~Cao, Y.~Bai, Z.~Zeng, M.~Ye, and M.~Zhang, ``An empirical study of clip for text-based person search,'' in \emph{Proceedings of the AAAI Conference on Artificial Intelligence}, vol.~38, no.~1, 2024, pp. 465--473.

\bibitem{ergasti2024mars}
A.~Ergasti, T.~Fontanini, C.~Ferrari, M.~Bertozzi, and A.~Prati, ``Mars: Paying more attention to visual attributes for text-based person search,'' \emph{arXiv preprint arXiv:2407.04287}, 2024.

\bibitem{qin2024noisy}
Y.~Qin, Y.~Chen, D.~Peng, X.~Peng, J.~T. Zhou, and P.~Hu, ``Noisy-correspondence learning for text-to-image person re-identification,'' in \emph{Proceedings of the IEEE/CVF Conference on Computer Vision and Pattern Recognition}, 2024, pp. 27\,197--27\,206.

\bibitem{zhang2023text}
S.~Zhang, D.~Cheng, W.~Luo, Y.~Xing, D.~Long, H.~Li, K.~Niu, G.~Liang, and Y.~Zhang, ``Text-based person search in full images via semantic-driven proposal generation,'' in \emph{Proceedings of the 4th International Workshop on Human-centric Multimedia Analysis}, 2023, pp. 5--14.

\bibitem{su2024maca}
L.~Su, R.~Quan, Z.~Qi, and J.~Qin, ``Maca: Memory-aided coarse-to-fine alignment for text-based person search,'' in \emph{Proceedings of the 47th International ACM SIGIR Conference on Research and Development in Information Retrieval}, 2024, pp. 2497--2501.

\bibitem{yan2025fusionsegreid}
J.~Yan, Y.~Wang, X.~Luo, and Y.-W. Tai, ``Fusionsegreid: Advancing person re-identification with multimodal retrieval and precise segmentation,'' \emph{arXiv preprint arXiv:2503.21595}, 2025.

\bibitem{zheng2015scalable}
L.~Zheng, L.~Shen, L.~Tian, S.~Wang, J.~Wang, and Q.~Tian, ``Scalable person re-identification: A benchmark,'' in \emph{Proceedings of the IEEE international conference on computer vision}, 2015, pp. 1116--1124.

\bibitem{li2014deepreid}
W.~Li, R.~Zhao, T.~Xiao, and X.~Wang, ``Deepreid: Deep filter pairing neural network for person re-identification,'' in \emph{Proceedings of the IEEE conference on computer vision and pattern recognition}, 2014, pp. 152--159.

\bibitem{zheng2017person}
L.~Zheng, H.~Zhang, S.~Sun, M.~Chandraker, Y.~Yang, and Q.~Tian, ``Person re-identification in the wild,'' in \emph{Proceedings of the IEEE conference on computer vision and pattern recognition}, 2017, pp. 1367--1376.

\bibitem{ding2021semantically}
Z.~Ding, C.~Ding, Z.~Shao, and D.~Tao, ``Semantically self-aligned network for text-to-image part-aware person re-identification,'' \emph{arXiv preprint arXiv:2107.12666}, 2021.

\bibitem{shu2022see}
X.~Shu, W.~Wen, H.~Wu, K.~Chen, Y.~Song, R.~Qiao, B.~Ren, and X.~Wang, ``See finer, see more: Implicit modality alignment for text-based person retrieval,'' in \emph{European Conference on Computer Vision}.\hskip 1em plus 0.5em minus 0.4em\relax Springer, 2022, pp. 624--641.

\bibitem{wang2024fine}
D.~Wang, F.~Yan, Y.~Wang, L.~Zhao, X.~Liang, H.~Zhong, and R.~Zhang, ``Fine-grained semantics-aware representation learning for text-based person retrieval,'' in \emph{Proceedings of the 2024 International Conference on Multimedia Retrieval}, 2024, pp. 92--100.

\bibitem{wei2024fine}
Z.~Wei, Z.~Zhang, P.~Wu, J.~Wang, P.~Wang, and Y.~Zhang, ``Fine-granularity alignment for text-based person retrieval via semantics-centric visual division,'' \emph{IEEE Transactions on Circuits and Systems for Video Technology}, 2024.

\bibitem{zhao2025cross}
W.~Zhao, Y.~Lu, Z.~Liu, Y.~Yang, and G.~Jiao, ``Cross-modal alignment with synthetic caption for text-based person search,'' \emph{International Journal of Multimedia Information Retrieval}, vol.~14, no.~2, p.~11, 2025.

\bibitem{chen2024sharegpt4v}
L.~Chen, J.~Li, X.~Dong, P.~Zhang, C.~He, J.~Wang, F.~Zhao, and D.~Lin, ``Sharegpt4v: Improving large multi-modal models with better captions,'' in \emph{European Conference on Computer Vision}.\hskip 1em plus 0.5em minus 0.4em\relax Springer, 2024, pp. 370--387.

\bibitem{chen2024expanding}
Z.~Chen, W.~Wang, Y.~Cao, Y.~Liu, Z.~Gao, E.~Cui, J.~Zhu, S.~Ye, H.~Tian, Z.~Liu \emph{et~al.}, ``Expanding performance boundaries of open-source multimodal models with model, data, and test-time scaling,'' \emph{arXiv preprint arXiv:2412.05271}, 2024.

\bibitem{bai2025qwen2}
S.~Bai, K.~Chen, X.~Liu, J.~Wang, W.~Ge, S.~Song, K.~Dang, P.~Wang, S.~Wang, J.~Tang \emph{et~al.}, ``Qwen2. 5-vl technical report,'' \emph{arXiv preprint arXiv:2502.13923}, 2025.

\bibitem{touvron2023llama}
H.~Touvron, T.~Lavril, G.~Izacard, X.~Martinet, M.-A. Lachaux, T.~Lacroix, B.~Rozi{\`e}re, N.~Goyal, E.~Hambro, F.~Azhar \emph{et~al.}, ``Llama: Open and efficient foundation language models,'' \emph{arXiv preprint arXiv:2302.13971}, 2023.

\bibitem{mitra2024compositional}
C.~Mitra, B.~Huang, T.~Darrell, and R.~Herzig, ``Compositional chain-of-thought prompting for large multimodal models,'' in \emph{Proceedings of the IEEE/CVF Conference on Computer Vision and Pattern Recognition}, 2024, pp. 14\,420--14\,431.

\bibitem{fan2024mllm}
J.~Fan, J.~Wu, J.~Gao, J.~Yu, Y.~Wang, H.~Chu, and B.~Gao, ``Mllm-sul: Multimodal large language model for semantic scene understanding and localization in traffic scenarios,'' \emph{arXiv preprint arXiv:2412.19406}, 2024.

\bibitem{guo2024regiongpt}
Q.~Guo, S.~De~Mello, H.~Yin, W.~Byeon, K.~C. Cheung, Y.~Yu, P.~Luo, and S.~Liu, ``Regiongpt: Towards region understanding vision language model,'' in \emph{Proceedings of the IEEE/CVF Conference on Computer Vision and Pattern Recognition}, 2024, pp. 13\,796--13\,806.

\bibitem{zhou2023regionblip}
Q.~Zhou, C.~Yu, S.~Zhang, S.~Wu, Z.~Wang, and F.~Wang, ``Regionblip: A unified multi-modal pre-training framework for holistic and regional comprehension,'' \emph{arXiv preprint arXiv:2308.02299}, 2023.

\bibitem{peng2023kosmos}
Z.~Peng, W.~Wang, L.~Dong, Y.~Hao, S.~Huang, S.~Ma, and F.~Wei, ``Kosmos-2: Grounding multimodal large language models to the world,'' \emph{arXiv preprint arXiv:2306.14824}, 2023.

\bibitem{chen2023shikra}
K.~Chen, Z.~Zhang, W.~Zeng, R.~Zhang, F.~Zhu, and R.~Zhao, ``Shikra: Unleashing multimodal llm's referential dialogue magic,'' \emph{arXiv preprint arXiv:2306.15195}, 2023.

\bibitem{ma2024groma}
C.~Ma, Y.~Jiang, J.~Wu, Z.~Yuan, and X.~Qi, ``Groma: Localized visual tokenization for grounding multimodal large language models,'' in \emph{European Conference on Computer Vision}.\hskip 1em plus 0.5em minus 0.4em\relax Springer, 2024, pp. 417--435.

\bibitem{rasheed2024glamm}
H.~Rasheed, M.~Maaz, S.~Shaji, A.~Shaker, S.~Khan, H.~Cholakkal, R.~M. Anwer, E.~Xing, M.-H. Yang, and F.~S. Khan, ``Glamm: Pixel grounding large multimodal model,'' in \emph{Proceedings of the IEEE/CVF Conference on Computer Vision and Pattern Recognition}, 2024, pp. 13\,009--13\,018.

\bibitem{niu2025chatreid}
K.~Niu, H.~Yu, M.~Zhao, T.~Fu, S.~Yi, W.~Lu, B.~Li, X.~Qian, and X.~Xue, ``Chatreid: Open-ended interactive person retrieval via hierarchical progressive tuning for vision language models,'' \emph{arXiv preprint arXiv:2502.19958}, 2025.

\bibitem{lei2107qvhighlights}
J.~Lei, T.~Berg, and M.~Bansal, ``Qvhighlights: Detecting moments and highlights in videos via natural language queries.(2021),'' \emph{URL https://arxiv. org/abs/2107.09609}.

\bibitem{mediapipe_pose}
``Mediapipe pose,'' \url{https://google.github.io/mediapipe/solutions/pose.html}, accessed: 2021-12-28.

\bibitem{zhang2024long}
B.~Zhang, P.~Zhang, X.~Dong, Y.~Zang, and J.~Wang, ``Long-clip: Unlocking the long-text capability of clip,'' in \emph{European Conference on Computer Vision}.\hskip 1em plus 0.5em minus 0.4em\relax Springer, 2024, pp. 310--325.

\bibitem{yin2025cross}
H.~Yin, X.~Man, F.~Chen, J.~Shao, and H.~T. Shen, ``Cross-modal full-mode fine-grained alignment for text-to-image person retrieval,'' \emph{arXiv preprint arXiv:2509.13754}, 2025.

\bibitem{wu2021application}
W.~Wu, H.~Liu, L.~Li, Y.~Long, X.~Wang, Z.~Wang, J.~Li, and Y.~Chang, ``Application of local fully convolutional neural network combined with yolo v5 algorithm in small target detection of remote sensing image,'' \emph{PloS one}, vol.~16, no.~10, p. e0259283, 2021.

\bibitem{wolf2019huggingface}
T.~Wolf, L.~Debut, V.~Sanh, J.~Chaumond, C.~Delangue, A.~Moi, P.~Cistac, T.~Rault, R.~Louf, M.~Funtowicz \emph{et~al.}, ``Huggingface's transformers: State-of-the-art natural language processing,'' \emph{arXiv preprint arXiv:1910.03771}, 2019.

\bibitem{yao2024minicpm}
Y.~Yao, T.~Yu, A.~Zhang, C.~Wang, J.~Cui, H.~Zhu, T.~Cai, H.~Li, W.~Zhao, Z.~He \emph{et~al.}, ``Minicpm-v: A gpt-4v level mllm on your phone,'' \emph{arXiv preprint arXiv:2408.01800}, 2024.

\end{thebibliography}
}

 





\end{document}